\documentclass[enabledeprecatedfontcommands, fontsize=10pt, a4paper, biblotoc]{scrartcl} 
\usepackage[english]{babel}
\usepackage[utf8]{inputenc}
\usepackage[svgnames]{xcolor}
\usepackage{amsmath, amssymb, amsthm}
\usepackage{mathrsfs}
\usepackage{geometry}
\usepackage{graphicx}
\usepackage{mathtools}
\usepackage{setspace}
\usepackage{enumerate}
\usepackage[automark]{scrlayer-scrpage}
\usepackage{esvect}
\usepackage{bibgerm}
\usepackage{dsfont}
\usepackage{pdfpages}
\usepackage{xparse}
\usepackage{todonotes}

\usepackage{chngcntr}

\usepackage{mathrsfs} 
\usepackage{caption}
\usepackage{pgf,tikz}
\usepackage{mathrsfs}
\usepackage{comment}

\allowdisplaybreaks

\usepackage{listings}
\usepackage{hyperref}
\usepackage[babel,german=quotes]{csquotes} 
\usepackage[style=authoryear-comp, natbib=true, backend=biber, doi=false, isbn=false, maxcitenames=2, uniquelist=false, dashed=false, firstinits=true]{biblatex}
\renewbibmacro{in:}{}
\DefineBibliographyExtras{english}{}
\renewbibmacro*{journal+issuetitle}{%
	\usebibmacro{journal}%
	\setunit*{\addcomma\space}%
	\iffieldundef{series}
	{}
	{\newunit
		\printfield{series}%
		\setunit{\addspace}}%
	\usebibmacro{volume+number+eid}%
	\setunit{\addspace}%
	\usebibmacro{issue+date}%
	\setunit{\addcolon\space}%
	\usebibmacro{issue}%
	\newunit}

\renewbibmacro*{volume+number+eid}{%
	\printfield{volume}%
	\setunit*{\addnbspace}
	\printfield{number}%
	\setunit{\addcomma\space}%
	\printfield{eid}}
\DeclareFieldFormat[article]{number}{\mkbibparens{#1}}
\DeclareNameAlias{sortname}{family-given}
\DeclareNameAlias{default}{family-given}

\usepackage{cleveref}
\usepackage[toc,page]{appendix}

\theoremstyle{thm} 
\newtheorem{theorem}{Theorem}[section]
\newtheorem{lemma}[theorem]{Lemma}
\newtheorem{proposition}[theorem]{Proposition}
\newtheorem{corollary}[theorem]{Corollary}

\theoremstyle{definition} 
\newtheorem{remark}[theorem]{Remark}
\newtheorem{definition}[theorem]{Definition}

\newtheorem{ass}[theorem]{Assumption}
\newtheorem{example}[theorem]{Example}
\numberwithin{equation}{section}

\automark{section}	
\lohead{}
\rohead[]{}
\ifoot[]{}
\ofoot[]{\pagemark}
\renewcommand{\phi}{\varphi}

\usepackage{bigints}
\usepackage{xfrac}

\newcommand{\A}{\mathcal{A}}
\newcommand{\D}{\mathrm{D}}
\newcommand{\Dr}{\mathds{D}}

\newcommand{\norm}[1]{\left\lVert#1\right\rVert}

\DeclareMathOperator*{\arginf}{arg\,inf}

\newcommand{\Risk}{\mathcal{R}}

\newcommand{\id}{\operatorname{id}}
\newcommand{\R}{\mathbb{R}}
\newcommand{\Rnp}{[0, \infty)}

\newcommand{\Cb}{\mathcal{C}_b}
\newcommand{\N}{\mathbb{N}}

\newcommand{\M}{\mathcal{M}}
\renewcommand{\L}{\mathscr{L}}

\newcommand{\W}{\mathcal{W}}
\newcommand{\Ld}{\mathscr{L}}

\newcommand{\email}[1]{\href{mailto:#1}{#1}}

\NewDocumentCommand\E{mg}{%
	\ensuremath{\mathbb{E}\IfNoValueF{#2}{_{#2}}\left[#1\right]}%
}

\makeatletter
\renewenvironment{proof}[1][\proofname] {\par\pushQED{\qed}\normalfont\topsep6\p@\@plus6\p@\relax\trivlist\item[\hskip\labelsep\bfseries#1\@addpunct{.}]\ignorespaces}{\popQED\endtrivlist\@endpefalse}
\makeatother
\newenvironment{Proof}[1]{\begin{proof}[Proof of #1]}{\end{proof}}

\newcommand{\Arg}[1]{\left(#1\right)}
\renewcommand{\epsilon}{\varepsilon}
\newcommand{\abs}[1]{\left\lvert #1 \right\rvert}

\newcommand*\diff{\mathop{}\!\mathrm{d}}

\renewcommand{\P}{\mathrm{P}}
\newcommand{\Qp}{\mathrm{Q}}

\renewcommand{\H}{\mathcal{H}}
\newcommand{\F}{\mathcal{F}}

\newcommand{\anf}[1]{``#1''}
\renewcommand{\L}{\mathcal{L}}
\newcommand{\sign}{\operatorname{sign}}
\newcommand{\X}{\mathcal{X}}
\newcommand{\Y}{\mathcal{Y}}
\newcommand{\Zc}{\mathcal{Z}}

\newlength{\leftstackrelawd}
\newlength{\leftstackrelbwd}
\def\leftstackrel#1#2{\settowidth{\leftstackrelawd}%
	{${{}^{#1}}$}\settowidth{\leftstackrelbwd}{$#2$}%
	\addtolength{\leftstackrelawd}{-\leftstackrelbwd}%
	\leavevmode\ifthenelse{\lengthtest{\leftstackrelawd>0pt}}%
	{\kern-.5\leftstackrelawd}{}\mathrel{\mathop{#2}\limits^{#1}}}

\DeclareMathOperator{\Exists}{\exists}
\DeclareMathOperator{\Forall}{\forall}

\def\smallskip{\vskip\smallskipamount}
\begin{document}\pagenumbering{roman} 
\begin{titlepage}
	\centering
\vspace{1.5cm}

{\huge\bfseries On the robustness of kernel-based pairwise learning\par}
\vspace{2cm}
{\LARGE\itshape Patrick Gensler~$^\text{1,}$\footnote[2]{~corresponding author Patrick Gensler, e-mail: \email{patrick.gensler@uni-bayreuth.de}}\\\vspace*{0.1cm}
	Andreas Christmann~$^\text{1}$\\\vspace*{0.3cm}}
{$^\text{1}$~Department of Mathematics, University of Bayreuth, Chair of Stochastics, 95440 Bayreuth, Germany}
\vspace*{4cm}

\textbf{Abstract~} It is shown that many results on the statistical robustness of kernel-based pairwise learning can be derived under basically no assumptions on the input and output spaces. In particular neither moment conditions on the conditional distribution of $Y$ given $X=x$ nor the boundedness of the output space is needed. We obtain results on the existence and boundedness of the influence function and show qualitative robustness of the kernel-based estimator. The present paper generalizes results by \citet{CZ2016} by allowing the prediction function to take two arguments and can thus be applied in a variety of situations such as ranking.
\vspace*{4cm}
$~~$\\
\today
	

\end{titlepage}




\section{Introduction}\pagenumbering{arabic}\setcounter{page}{1}

\citet{stute1991, stute1994} showed the (universal) consistency of conditional U-statistics under weak conditions. Based on these results, \citet{CLV2008} reused U-statistics in the field of statistical learning theory and more precisely for the ranking problem. The present paper is connected with these papers by studying the \emph{application of U-statistics as potential loss functions} in the field of statistical machine learning. \\

As mentioned above, an example are ranking based problems, that can be simplified to a situation where profiles of entities are given and have to be compared against each other to find the order of these entities in a particular case. For instance an employer is interested in two candidates and  wants to select the \anf{better} one for the company considering their applications and the experience of the employer from former employees. The same problem in another setting can be found in a lot of fields such as insurance companies, banks, product marketing, etc..\\

In the field of statistical machine learning theory, one approach are kernel-based methods such as support vector machines, see e.\,g. \citet{vapnik1995, vapnik1998} for classical textbooks on this subject. The field of kernel-based learning methods has been widely researched, refer to for instance \citet{christiani2000}, \citet{schoelkopf2001},  \citet{cucker2007} and \citet{book}.
Kernel-based pairwise learning methods were studied by e.\,g. \citet{CZ2016}. They showed the statistical robustness of pairwise learning methods in the sense of bounded influence functions and qualitative robustness, as introduced by \citet{hampel1971}, \citet{hampel1986} and generalized by \citet{cuevas1988}. The difference compared to classic support vector machines is that the loss function does not only take three arguments $(x,y,f(x))$, where $x$ from the input space $\X$, $y$ from the output space $\Y$ and the prediction $f(x)$, but rather six $x,y,x',y'$ as pairwise components - what explains the modified name pairwise loss functions - and the predictions $f(x)$ and $f(x')$. \\

In this article, we analyze several statistical robustness properties of kernel-based pairwise learning methods based on pairwise loss functions $L(x,y,x',y',f(x,x'))$ that take five arguments with the real valued prediction function taking two arguments $f(x,x')$. This is an additional generalization compared to \citet{CZ2016} as the difference $\tilde{f}(x)-\tilde{f}(x') = f(x,x')$ investigated in the mentioned paper is a special case for a \emph{\anf{bivariate} predicition function} with $\tilde{f}$ being a suitable function.\\

\citet[p.\,1375]{rejchel2012} investigated the ranking problem in a similar way, but considered a uniformly bounded function class with $f(x,x') = -f(x',x)$ or parametrized ranking rules $f(x,x') = \theta^T(x-x')$ in the regression setting with $\X = \R^d, \Y = \R$ and parameter $\theta \in \R^d$ (\cite*[p.\,6]{Rej2017}). In the context of online learning, we refer to \citet{ying2015} and \citet{guo2017} and for metric learning to \citet{bellet2015} and \citet{cao2016} and the references cited therein. \\ 

Using \emph{shifted loss functions}, see \Cref{shifted-loss}, to tackle the robustness problem for support vector machines in the case of heavy-tailed distributions was already done by  \citet{CvMS2009} and is applied here in the context of regularized pairwise learning in order to be able to compute prediction functions without any moment assumption on the output variable $Y$. To be more precise, shifted loss functions are a technical tool to avoid moment conditions of $Y$ given $X=x$ without changing the estimator, if the estimator exists based on the unshifted loss function. \\

The paper is organized as follows. \Cref{mathPrereq.} introduces the necessary mathematical prerequisites. Readers familiar with kernel-based pairwise learning can skip this section. \Cref{mainResults} presents the main results: a general representer theorem, the risk consistency, and the robustness of the kernel-based regularized pairwise learning method. \Cref{discussion} gives a discussion and an outlook for further research topics in this field. Additional theorems and lemmas, as well as proofs for our results are listed in the \Cref{appendix-1} or \Cref{appendix-2}, respectively.

\section{Mathematical prerequisites}\label{mathPrereq.}

In this section we collect some definitions and results which are useful to study regularized pairwise learning. If not mentioned otherwise, we enclip topological spaces $(\mathcal{Z}, \tau_\mathcal{Z})$ with their Borel-$\sigma$-algebras $\mathcal{B}(\mathcal{Z})$. For brevity we denote the set of all Borel probability measures on a topological space $(\mathcal{Z}, \tau_\mathcal{Z})$ as $\M_1(\mathcal{Z})$ instead of $\M_1(\mathcal{Z}, \mathcal{B}(\mathcal{Z}))$. We denote the set of all continuous bounded function $f: \Zc \to \R$ by $\Cb(\Zc)$.\\

As mentioned in the introduction, let $\X$, the input space, and $\Y$, the output space, be topological spaces.\\

We define the set of measurable functions $f: \Zc \to \R$ with $\L_0(\Zc)$. The set of all measurable functions $f: \X \times \Y \to \R$ satisfying $\int f \diff \P < \infty$ is defined by $\L_1(\X \times \Y, \P)$ (or short $\L_1(\P)$, if the domain is obvious from the context) with $\P \in \M_1(\X \times \Y)$. For all measurable functions $f: \X \times \Y \to \R$ that are almost surely bounded, given a probability measure $\P \in \M_1(\X \times \Y)$, we write $\L_\infty(\X \times \Y, \P)$ (or short $\L_\infty(\P)$). 

A kernel $k: \X^2 \times \X^2 \to \R$ is called \textbf{bounded} if $\norm{k}_\infty := \sup_{(x, x') \in \X^2} \sqrt{k((x,x'),(x,x'))} < \infty$.
We refer to \citet{berlinet2004} or \citet[Chapter 4]{book} for a thorough introduction to kernels and reproducing kernel Hilbert spaces. See also \Cref{appendix-1}.

\begin{definition}
	Let $(\X,\A)$ be a measurable space and $\Y \subset \R$ be a closed subset. Then a measurable function $L: (\X \times \Y)^2 \times \R \to \Rnp$ is called a \textbf{pairwise loss function} or \textbf{pairwise loss} in short.
\end{definition}

\begin{example}\label{CLossExp}
	An example of a pairwise loss function is given by \citet[p.\,863]{CLV2008} utilizing an auxiliary measurable function $\phi: \R \to \Rnp$ satisfying the following two conditions: $\phi(0) = 1$ and $\phi(x) \geq 1$ for all $x \in \Rnp$.
It is then possible to define the loss function 
	\begin{align*}
	L(x,y,x',y',f(x,x')) :&= \phi\Arg{\sign\Arg{y-y'}f(x,x')},
	\end{align*}
	with the sign function at $0$ being defined as $\sign(0)=0$ and $\phi$ chosen as for instance the exponential function $\exp(x)$, the function $\log_2(1+\exp(x))$ or a hinge loss with $\max\{0, 1+x\}$.
	Replacing the sign function by a differentiable surrogate function leads to the following \anf{smoothed} pairwise loss function
	\begin{align*}
	L_\sigma(x,y,x',y',f(x,x')) := \phi\Arg{\tanh\Arg{\sigma^{-1}(y-y')}f(x,x')},
	\end{align*}
	with an arbitrary small $\sigma > 0$.
\end{example}

\begin{example}
	Another example for the loss function is the \textbf{least squares ranking loss} used by \citet[p.\,55]{chen2014}
	\begin{align*}
	L(x,y,x',y',f(x,x')) := (y-y'-f(x,x'))^2,
	\end{align*}
	with $f(x,x') := \tilde{f}(x) - \tilde{f}(x')$ for a univariate prediction function $\tilde{f}: \X \to \R$.
\end{example}

We now define several quantities we will need to introduce our kernel-based pairwise learning.

\begin{definition}
	Let $L: (\X \times \Y)^2 \times \R \to \Rnp$ be a pairwise loss function, $\P \in \M_1(\X \times \Y)$, and $\P^2 = \P \otimes \P$ denoting the product measure of $\P$. 
	\begin{enumerate}[(a)]
		\item Then, for a measurable function $f: \X^2 \to \R$, the \textbf{$L$-risk} is defined by 
		\begin{align*}
		\Risk_{L,\P}(f) &:= \E{L(X,Y,X',Y',f(X,X'))}{\P^2} =\int_{(\X \times \Y)^2} L(x,y,x',y',f(x,x')) \diff \P^2(x,y,x',y').
		\end{align*}
		\item The minimal L-risk 
		\begin{align*}
		\Risk_{L,\P}^* := \inf_{f \in \L_0(\X^2)} \Risk_{L,P}(f)
		\end{align*}
		is called the \textbf{Bayes risk} and a measurable minimizer $f_{L, \P}: \X^2 \to \R$ is called a \textbf{Bayes decision function}, if it exists.
	\end{enumerate}

\end{definition}

\begin{remark}
	If $(\X, \tau)$ is a Polish space (with topology $\tau$) and $\Y \subset \R$ is closed, then $\X \times \Y$ is a Polish space and so is $(\X \times \Y)^2$ as a countable product of Polish spaces, see e.g. \citet[p.\,13]{kechris1995}. Hence we can split up $\P$ into the conditional probability of $Y$ given $X$ and the marginal distribution $\P_X$, i.e.
	\begin{align*}
	\Risk_{L,\P}(f) &=\int_{(\X \times \Y)^2} L(x,y,x',y',f(x,x')) \diff \P^2(x,y,x',y')\\
	&=\int_\X\int_\Y\int_\X\int_\Y L(x,y,x',y',f(x,x')) \P(\diff y|x)\P_X(\diff x)\P(\diff y'|x')\P_X(\diff x'),
	\end{align*}
	see \citet[Sect.~10.2]{dudley2002}.
\end{remark}

Computing the infimum of the risk over the set of all measurable functions for empirical distributions $\D$ instead of $\P$ is in general not doable and might lead to overfitting. In order to reduce the danger of overfitting, one approach is to introduce a regularizing term to penalize such estimated predictor functions. Another modification that can be made is to restrict the set that the risk is minimized over from all measurable functions to a reproducing kernel Hilbert space (RKHS) $\H$ of a measurable kernel $k: \X^2 \times \X^2 \to \R$ in order to simplify the computation. If a universal kernel, such as the Gaussian RBF or the Laplacian kernel, is chosen, then every continuous prediction function can be arbitrarily approximated due to the denseness of the corresponding RKHS in the space of continuous functions (see e.g. \citet[p.\,152ff.]{book}). Both ways are used in the setting of support vector machines and regularized pairwise learning.

The remarks above lead to the introduction of a regularized version of the risk.

\begin{definition}\label{regRisk}
		Let $L: (\X \times \Y)^2 \times \R \to \Rnp$ be a pairwise loss function and $\P \in \M_1(\X \times \Y)$. Then, for $f \in \L_0(\X^2)$ and $\lambda > 0$, the \textbf{regularized $L$-risk} is defined by 
		\begin{align*}
		\Risk_{L,\P, \lambda}^\text{reg}(f) := \Risk_{L,\P}(f) + \lambda\norm{f}_\H^2.
		\end{align*}
		The corresponding minimizer is then abbreviated with $f_{L, \P, \lambda}: \X^2 \to \R$,
		\begin{align*}
		f_{L, \P, \lambda} = \arginf_{f \in \H} \Risk_{L,\P, \lambda}^\text{reg}(f).
		\end{align*}
\end{definition}

The following definitions, theorems and lemmas are taken from \citet{CvMS2009} and \citet{CZ2016}.

\begin{lemma}\label{lem:MeasurabilityLossRisk}
	Let $L$ be a pairwise loss function and $\F \subset \L_0(\X^2)$ be a subset that is equipped with a complete and separable metric $d$ and its corresponding Borel-$\sigma$-algebra. Assume that the metric $d$ dominates pointwise convergence, i.e. 
	\begin{align*}
	\lim_{n \to \infty} d(f_n,f) = 0 \Longrightarrow \lim_{n \to \infty} f_n(x,x') = f(x,x') ~\Forall (x,x') \in \X^2,\,\Forall f, f_n \in \F.
	\end{align*}
	Then the evaluation map $\F \times \X^2 \to \R$ defined by $(f,(x,x')) \mapsto f(x,x')$ is measurable and consequently the map $(x,y,x',y',f) \mapsto L(x,y,x',y',f(x,x'))$ defined on $(\X \times \Y)^2 \times \F$ are also measurable. Finally given $\P \in \M_1(\X\times \Y)$, the risk functional $\Risk_{L,P}: \F \to \Rnp$ is measurable.
\end{lemma}

\begin{definition}
	A pairwise loss function $L$ is called
	\begin{enumerate}[(i)]
		\item (\textbf{strictly}) \textbf{convex}, \textbf{continuous} or \textbf{differentiable}, if $L(x,y,x',y',\cdot): \R \to \Rnp$
		is (strictly) convex, continuous or differentiable for all $(x,y,x',y') \in (\X \times \Y)^2$, respectiviely. Denote the partial Fréchet derivative with respect to the fifth argument by $D_5L$.
		\item \textbf{locally Lipschitz continuous}, if, for all $b \geq 0$, there exists a constant $c_b \geq 0$ such that, for all $t,t' \in [-b,b]$, we have
		\begin{align*}
		\sup_{\substack{x,x' \in \X\\y,y' \in \Y}} \abs{L(x,y,x',y',t) - L(x,y,x',y',t')} \leq c_b\abs{t-t'}. \hfill \tag{$\ast$}
		\end{align*}
		Moreover, for $b \geq 0$, the smallest such constant $c_b$ is denoted by $\abs{L}_{b,1}$. Furthermore, $L$ is called \textbf{Lipschitz continuous}, if there exists a constant $\abs{L}_1 \in \Rnp$ such that, for all $t,t' \in \R$, the supremum in $(\ast)$ is less than or equal to $\abs{L}_1 \cdot \abs{t-t'}$.
	\end{enumerate}
\end{definition}

\begin{example}
	The loss function from \Cref{CLossExp} is Lipschitz continuous and convex, if the auxiliary function $\phi$ is as well. Let $t_1, t_2 \in \R$ and $\abs{\phi}_1$ the Lipschitz constant of $\phi$
	\begin{align*} 
	\abs{L(x,y,x',y',t_1) - L(x,y,x',y',t_2)} &= \abs{\phi\Arg{\sign\Arg{y-y'}t_1} - \phi\Arg{\sign\Arg{y-y'}t_2}}\\
	&\leq \abs{\phi}_1\abs{\sign\Arg{y-y'}}\abs{t_1-t_2}\\
	& \leq \abs{\phi}_1\abs{t_1-t_2}.
	\end{align*}
	Choosing for instance $\phi(x) = \log(1+e^x)$, then the Lipschitz continuity follows from the boundedness of its derivative $\phi'(x) = \frac{e^x}{1+e^x}$. For $\phi$ the exponential function we can only yield local Lipschitz continuity.  
	
	For convex $\phi$ it follows with $\upsilon \in [0,1]$
	\begin{align*}
	L(x,y,x',y',\upsilon t_1 + (1-\upsilon)t_2) &= \phi\Arg{\sign\Arg{y-y'}(\upsilon t_1 + (1-\upsilon)t_2)}\\
	&= \upsilon\phi\Arg{\sign\Arg{y-y'}t_1} - (1-\upsilon)\phi\Arg{\sign\Arg{y-y'}t_2}\\
	&= \upsilon L(x,y,x',y',\upsilon t_1) + (1-\upsilon)L(x,y,x',y',(1-\upsilon)t_2).
	\end{align*}
	Both properties are also satisfied by the smoothed version in \Cref{CLossExp}. Differentiability with respect to the fifth argument is guaranteed by both examples and only depends on the differentiability of the auxiliary function.
\end{example}

\begin{lemma}\label{RiskConvex}
	Let $L$ be a (strictly) convex loss function and $\P \in \M_1(\X \times \Y)$. Then $\Risk: \L_0(\X^2) \to [0, \infty]$ is (strictly) convex.
\end{lemma}

\begin{lemma}\label{RiskConti}
	Let $\P \in \M_1(\X \times \Y)$ and $L$ be a locally Lipschitz continuous pairwise loss function. Then for all $B \geq 0$ and all $f,g \in \L_\infty(\P_X^2)$ with $\norm{f}_\infty \leq B$ and $\norm{g}_\infty \leq B$, we have
	\begin{align*}
	\abs{\Risk_{L,\P}(f)-\Risk_{L,\P}(g)} \leq \abs{L}_{B,1}\norm{f-g}_{L_1(\P_X^2)}.
	\end{align*}
	Furthermore, the risk functional $\Risk_{L,\P}: \L_\infty(\P_X^2) \to \Rnp$ is well-defined and continuous.
\end{lemma}

	For $f,g \in \H$ and a Lipschitz continuous loss function $L$ the lemma yields
	\begin{align*}
		\abs{\Risk_{L,\P}(f)-\Risk_{L,\P}(g)} &\leq \abs{L}_1\norm{f-g}_{\L_1(\P_X^2)} \leq \abs{L}_1\norm{f-g}_\infty \leftstackrel{}{\leq} \ \abs{L}_1\norm{k}_\infty\norm{f-g}_\H.
	\end{align*}

\begin{definition}
	A pairwise loss function $L: (\X \times \Y)^2 \times \R \to \Rnp$ is called a \textbf{pairwise Nemitski loss function} if a measurable function $b: (\X \times \Y)^2 \to \Rnp$ and a monotonically increasing function $h: \Rnp \to \Rnp$ exist, such that
	\begin{align*}
	L(x,y,x',y',t) \leq b(x,y,x',y') + h(\abs{t}), &\qquad (x,y,x',y',t) \in (\X \times \Y)^2 \times \R.
	\end{align*}
\end{definition}

\begin{lemma}\label{Risk-dbar}
	Let $\P \in \M_1(\X \times \Y)$ and $L$ be a differentiable pairwise loss function such that $\abs{D_5L}$ is a $\P$-integrable Nemitski loss function. Then the risk function $\Risk_{L,\P}: \L_\infty(\P_X^2) \to \Rnp$ is Fréchet differentiable and its derivative at $f \in \L_\infty(\P_X^2)$ is the bounded linear operator $\Risk'_{L,\P} : \L_\infty(\P_X^2) \to \R$ with 
	\begin{align*}
	\Risk'_{L,\P}(f)g = \int_{(\X \times \Y)^2} D_5L(x,y,x',y',f(x,x'))g(x,x') \diff \P^2(x,y,x',y').
	\end{align*}
\end{lemma}

	If in \Cref{Risk-dbar} the derivative of the pairwise loss function with respect to the fifth argument is continuous and uniformly bounded for all $x, x' \in \X$ and all $y, y' \in \Y$ by a constant $c_L \in \Rnp$, then upper assertion follows immediately from the lemma because
	\begin{align*}
	\abs{D_5L(x,y,x',y',t)} \leq c_L, \quad \Forall (x,y,x',y',t) \in (\X \times \Y)^2 \times \R
	\end{align*}
	and thus the condition that $\abs{D_5L}$ is a $\P^2$-integrable Nemitski loss function follows, because we can set $b(x,y,x',y') \equiv c_L$ and $h(\abs{t}) \equiv 0$ or the other way around.\\

The following introduction of shifted loss functions offers more possibilites to determine risks and hence a risk-minimizing function in the pairwise learning setting due to it not being dependent on the conditional distribution of $Y$ given $X = x$. Although the definitions above demand a measurable pairwise loss function and therefore especially a non-negative function, they can also be used in a more generalized situation.

\begin{definition}\label{shifted-loss}
	Let $L$ be a pairwise loss function, the corresponding \textbf{shifted pairwise loss function} $L^*: (\X \times \Y)^2 \times \R \to \R$ is defined by
	\begin{align*}
	L^*(x,y,x',y',t) := L(x,y,x',y',t) - L(x,y,x',y',0).
	\end{align*}
\end{definition}

\begin{remark}
	By using the shifted loss function of a Lipschitz continuous pairwise loss it is possible to make the risk \emph{independent} of the moments of $Y$
	\begin{align*}
	\abs{\Risk_{L^*,\P}(f)} &\leq \E{\abs{L(X,Y,X',Y',f(X,X')) - L(X,Y,X',Y',0)}}{\P^2}\\
	&\leq \E{\abs{L}_1\abs{f(X,X')-0}}{\P^2} = \abs{L}_1 \E{\abs{f(X,X')}}{\P^2}\\
	&\leq \abs{L}_1 \norm{f}_{\L_\infty(\X^2, \P_X^2)} < \infty,
	\end{align*}
	with $\P_X$ denoting the marginal distribution of $X$. This can be guaranteed by for instance choosing a measurable and bounded kernel $k: \X^2 \times \X^2 \to \R$ and $f \in \H$ with $\H$ being $k$'s corresponding reproducing kernel Hilbert space.
\end{remark}

We will now prove several lemmas to be able to show the uniqueness and existence of (regularized) Risk-minimizing functions in the pairwise learning setting.

\begin{lemma}\label{shifted-convex}
	Let $L$ be a pairwise loss function. Then the following statements concerning the corresponding shifted loss function $L^*$ are valid.
	\begin{enumerate}[(i)]
		\item $L^*$ is (strictly) convex, if $L$ is (strictly) convex.
		\item $L^*$ is Lipschitz continuous, if $L$ is Lipschitz continuous. Furthermore, both Lipschitz constants are equal, i.e. $\abs{L}_1 = \abs{L^*}_1$.
	\end{enumerate}
\end{lemma}

	A shifted Lipschitz continuous pairwise loss function is a Nemitski loss function, because it follows
	\begin{align*}
	\abs{L^*(x,y,x',y',t)} &= \abs{L(x,y,x',y',t) - L(x,y,x',y',0)} \leq \abs{L}_1 \abs{t}
	\end{align*}
	and thus the property of a Nemitski loss function with $b(x,y,x',y') := \abs{L}_1$ and $h(\abs{t}) = \abs{t}$. If $f \in \L_1(\P_X^2)$, then $L^*$  is a $\P^2$-integrable Nemitski loss function with $t \equiv f(x,x')$.

\begin{lemma}\label{Inequalities}
	The following assertions are valid for shifted pairwise loss functions $L^*$.
	\begin{enumerate}[(i)]
		\item \label{inequality1} \begin{equation}
			\begin{aligned}\inf\limits_{t \in \R} L^*(x, y, x',y',t) \leq 0.\end{aligned}
		\end{equation}
		\item \label{inequality2} If $L$ is a Lipschitz continuous loss function, then for all $f \in \H$
		\begin{align}
		-\abs{L}_1\E{\abs{f(X,X')}}{\P^2} \leq \Risk_{L^*,\P}(f) &\leq \abs{L}_1\E{\abs{f(X,X')}}{\P^2},\\
		-\abs{L}_1\E{\abs{f(X,X')}}{\P^2} + \lambda\norm{f}_\H^2 \leq \Risk^\text{reg}_{L^*,\P, \lambda}(f) &\leq \abs{L}_1\E{\abs{f(X,X')}}{\P^2}+\lambda\norm{f}_\H^2,
		\end{align}
		\item \label{inequality3} $\inf\limits_{f \in \H} \Risk^\text{reg}_{L^*,\P, \lambda}(f) \leq 0$ and hence $\inf\limits_{f \in \H} \Risk_{L^*,\P}(f) \leq 0.$
		\item \label{inequality4} Let $L$ be a Lipschitz continuous loss function and assume that $f_{L^*, \P, \lambda}$ exists. Then we have
		\begin{align}
		\lambda\norm{f_{L^*, \P, \lambda}}_\H^2 &\leq -\Risk_{L^*,\P}(f_{L^*, \P, \lambda}) \leq \Risk_{L,\P}(0),\\
		0 &\leq -\Risk^\text{reg}_{L^*,\P, \lambda}(f) \leq \Risk_{L,\P}(0),\\
		\lambda\norm{f_{L^*, \P, \lambda}}_\H^2 &\leq \min\{\abs{L}_1\E{\abs{f_{L^*,\P,\lambda}}}{\P^2}\}, \Risk_{L,\P}(0)\}.
		\end{align}
		If the kernel $k: \X^2 \times \X^2 \to \R$ is additionally bounded, then
		\begin{align}
		\norm{f_{L^*, \P, \lambda}}_\infty &\leq \lambda^{-1}\abs{L}_1\norm{k}_\infty^2 < \infty,\\
		\abs{\Risk_{L^*,\P}(f_{L^*,\P,\lambda})} &\leq \lambda^{-1}\abs{L}^2_1\norm{k}_\infty^2 < \infty.
		\end{align}
		\item \label{inequality5} If the partial Fréchet derivatives of $L$ and $L^*$ exist for $(x,y,x',y') \in (\X \times \Y)^2$, then
		\begin{align}
		D_5 L^*(x,y,x',y',t) &= D_5 L(x,y,x',y',t), \qquad \Forall t \in \R.
		\end{align}
	\end{enumerate}
\end{lemma}

\begin{lemma}\label{LipschitzRisk}
	Let $L$ be a Lipschitz continuous pairwise loss and $f \in \L_1(\X^2, \P_X^2)$. Then $\Risk_{L^*,\P}(f) \notin \{-\infty, \infty\}$. Moreover, we have $\Risk^\text{reg}_{L^*,\P, \lambda}(f) > -\infty$ for all $f \in \L_1(\X^2, \P_X^2) \cap \H$.
\end{lemma}

A \textbf{regularized pairwise learning method} is an operator which maps probability measures $\P \in \M_1(\X \times \Y)$ to a corresponding regularized risk minimizing function $f$ in the RKHS $\H$,  $S: \M_1(\X \times \Y) \to \H: \P \mapsto S(\P) = f_{L^*, \P, \lambda}$ for any given $\lambda > 0$. \\

In the following assertions the classic problem of existence and uniqueness of such minimizers is taken care of.

\begin{theorem}[Uniqueness of minimizer]\label{thm:uniquenessMinimizer}
	Let $L$ be a convex pairwise loss function. Assume that 
	\begin{enumerate}[(i)]
		\item $\Risk_{L^*,\P}(f) < \infty$ for some $f \in \H$ and $\Risk_{L^*,\P}(f) > -\infty$ for all $f \in \H$
	\end{enumerate}
	or
	\begin{enumerate}[(i)]
		\setcounter{enumi}{1}
		\item $L$ is Lipschitz continuous and $f \in \L_1(\P_X^2)$ for all $f \in \H$.
	\end{enumerate}
	Then, for all $\lambda > 0$, there exists at most one solution $f_{L^*,\P,\lambda}$.
\end{theorem}

\begin{theorem}[Existence of minimizer]\label{thm:existenceMinimizer}
	Let $L$ be a Lipschitz continuous, convex pairwise loss function and $\H$ be the RKHS of a bounded measurable kernel $k$. Then, for all $\lambda > 0$, there exists a minimizing prediction function $f_{L^*,\P,\lambda}$.
\end{theorem}

The theorem above shows the existence of a Bayes decision function for regularized pairwise learning methods, if the loss function is convex. The Minimum Error Entropy (MEE) loss function, see e.g. \citet[p.\,5f.]{CZ2016}, is a leading example for a non-convex loss function and therefore it is relevant to prove the existence of a minimizer in such a case as well. 

\begin{theorem}\label{thm:existenceMinimizerNonConvex}
	If $L$ is a Lipschitz continuous pairwise loss function, $\P\in \M_1(\X \times \Y), \Risk_{L,\P}(f_0) < \infty$ for some $f_0 \in \H$, and $\H$ the RKHS of a bounded and measurable kernel $k$ on $\X^2$, then a minimizer $f_{L, \P, \lambda} \in \H$ exists for any $\lambda > 0$.
\end{theorem}

For a general representer theorem, a couple of notational remarks and the introduction of the subdifferential are necessary.\\

Let $E$ be a Banach space, $E'$ its dual space, and $x \in E, x' \in E'$. A common notation is the so-called \textbf{dual pairing} $\left<x',x\right>_{E', E} := x'(x).$

Let $f: E \to \R \cup \{\infty\}$ be a convex function and $w \in E$ with $f(w) < \infty$. Then the \textbf{subdifferential} of $f$ at $w$ is defined by
\begin{align*}
\partial f(w) :&= \{w' \in E': \left<w', v-w\right>_{E',E} \leq f(v) - f(w) \quad \Forall v \in E \}\\
&= \{w' \in E': w'(v-w) \leq f(v) - f(w) \quad \Forall v \in E \}.
\end{align*}

The main result in this section is the following representer theorem which can eventually be proven with the same methods as in the case of support vector machines.

\begin{theorem}[Representer Theorem]\label{representerTheorem}
	Let $L$ be a convex and Lipschitz continuous pairwise loss function, $L^*$ its corresponding shifted loss function, $k$ be a bounded and measurable kernel with separable RKHS $\H$. Then, for all $\lambda > 0$, there exists an $h_\P \in \L_\infty((\X \times \Y)^2, \P^2)$ such that
	\begin{enumerate}[(i)]
	\item $\begin{aligned}[t] 
	h_\P(x,y,x',y') &\in \partial L^*(x,y,x',y',f_{L^*,\P,\lambda}(x,x')), \quad \Forall (x,y,x',y') \in (\X \times \Y)^2
	\end{aligned}$
	\item $\begin{aligned}[t]
	f_{L^*,\P,\lambda} &= -(2\lambda)^{-1}\E{h_\P \Phi}{\P^2},
	\end{aligned}$
	\item $\begin{aligned}[t]
		\norm{h_\P}_\infty &\leq \abs{L^*}_1,
	\end{aligned}$
	\item $\begin{aligned}[t]
	\norm{f_{L^*,\P,\lambda} - f_{L^*,\Qp,\lambda}}_\H &\leq \lambda^{-1}\norm{\E{h_\P \Phi}{\P^2} - \E{h_\P \Phi}{\Qp^2}}_\H, \quad  \forall \, \Qp \in \M_1(\X \times \Y).
	\end{aligned}$
	\end{enumerate}
\end{theorem}

\section{Main Results}\label{mainResults}

In this section we will give our main results: consistency and (qualitative) robustness for the regularized pairwise learning method. The proofs are given in the \Cref{appendix-2}. We use techniques from \citet{CvMS2009} and \citet{CZ2016}. However, here we treat the more general case of prediction functions $f: \X \times \X \to \R$ instead of the well-investigated case $f: \X \to \R$ by the authors mentioned above. \\

The first result is that the risk of the empirical prediction function converges to the Bayes risk under regularity assumptions.

\begin{theorem}[Risk consistency]\label{riskConsistency}
	Let $L$ be a convex, Lipschitz continuous pairwise loss function, $L^*$ the corresponding shifted version and $\H$ be a separable RKHS of a bounded measurable kernel $k: \X^2 \times \X^2 \to \R$ such that $\H$ is dense in $\L_1(\X^2, \mu)$ for all $\mu \in \M_1(\X^2)$. Let $(\lambda_n)_{n \in \N} \subset (0, \infty)$ be a sequence with $\lambda_n \to 0$.
	\begin{enumerate}[(i)]
		\item If $\lambda_n^2n \to \infty$, then for all $\P \in \M_1(\X \times \Y)$, $\Risk_{L^*,\P}(f_{L^*,\D,\lambda_n}) \to \Risk^*_{L^*,\P}$ in probability for $n \to \infty$ and all sets of data with $|D|=n$.
		\item If $\lambda_n^{2+\delta}n \to \infty$ for some $\delta > 0$, then the convergence above holds $\P$-almost surely.
	\end{enumerate}
\end{theorem}

The next result gives an upper bound for the $\H$-norm of the difference between minimizers of the probability measure $\P$ and a contaminated probability measure $\P_\epsilon$, which is a mixture of $\P$ and another probability measure $\Qp$.

\begin{theorem}[Bounds for bias]\label{maxBias}
	Let $\H$ be a separable RKHS of a bounded and measurable kernel $k: \X^2 \times \X^2 \to \R$. Then, for all $\lambda > 0$, all $\epsilon \in (0,1)$, and all probability measures $\P, \Qp \in \M_1(\X \times \Y)$, we have, for all $\P_\epsilon = (1-\epsilon)\P + \epsilon \Qp \in \M_1(\X \times \Y)$,
	\begin{align*}
	\norm{f_{L^*,\P,\lambda} - f_{L^*,\P_\epsilon, \lambda}}_\H \leq c_{\P,\Qp}\epsilon,
	\end{align*}
	where $c_{\P,\Qp} = \frac{8}{\lambda}\norm{k}_\infty \abs{L}_1$.
\end{theorem}

For our results on the statistical robustness, we require the following technical assumptions.

\begin{ass}\label{assumptions}
	Let the following assumptions be satisfied in this section.
	\begin{enumerate}[(i)]
		\item Let $\Y \subset \R$ be a closed subset and $\X$ a complete separable metric space. Let $(X_i, Y_i, X'_i, Y'_i)$ be a tuple of $(\X \times \Y)^2$-valued random elements, which are independent and identically distributed with $\P \in \M_1(\X \times \Y)$ being the distribution of $(X_i, Y_i)$ and $(X_i', Y_i')$.
		\item Let $k: \X^2 \times \X^2 \to \R$ be a continuous and bounded kernel with separable RKHS $\H$, see e.g. \citet[Cor.~4, p.\,35]{berlinet2004}, and $\Phi: \X^2 \to \H: \Phi(x,x') := k\bigl((\cdot, \cdot), (x, x')\bigr)$ with $(x,x') \in \X^2$ being the canonical feature map.
		\item 	Let $L$ be a Lipschitz continuous, convex, differentiable pairwise loss function for which the first and second partial derivatives with respect to the last argument are continuous and bounded
		\begin{itemize}
			\item $\sup\limits_{\substack{x,x' \in \X\\y,y' \in \Y}} \abs{D_5L(x,y,x',y',\cdot)} \leq c_{L,1} \in (0, \infty)$
			\item $\sup\limits_{\substack{x,x' \in \X\\y,y' \in \Y}} \abs{D_5D_5L(x,y,x',y',\cdot)} \leq c_{L,2} \in (0, \infty)$.
		\end{itemize}
	\end{enumerate}
\end{ass}

\begin{theorem}\label{RPLGateaux}
	For all Borel probability measures $\P, \Qp \in \M_1(\X \times \Y)$, the \textbf{regularized pairwise learning} operator (\textbf{RPL} operator).
	\begin{align*}
	S: \M_1(\X \times \Y) \to \H, \qquad S(P) := f_{L^*,\P, \lambda},
	\end{align*}
	has a bounded Gâteaux derivative $S'_G(\P)$ at $\P$ and
	\begin{align*}
	S_G'(\P)(\Qp) = -M(\P)^{-1}T(\Qp;\P).
	\end{align*}
	To shorten the notation, we write $L'_{f_{L^*, \P, \lambda}}(X,Y,X',Y') := D_5L(X, Y, X', Y', f_{L^*, \P, \lambda}(X, X'))$. Then,
	\begin{align*}
	T(\Qp;\P) &= -2\E{L'_{f_{L^*, \P, \lambda}}(X,Y,X',Y')\Phi(X,X')}{\P^2} + \E{L'_{f_{L^*, \P, \lambda}}(X,Y,X',Y')\Phi(X,X')}{\P \otimes \Qp}  + \\
	&\qquad \E{L'_{f_{L^*, \P, \lambda}}(X,Y,X',Y')\Phi(X,X')}{\Qp \otimes \P}
	\end{align*}
	equals the gradient of the regularized risk and
	\begin{align*}
	M(P) = 2\lambda\id_\H + \E{D_5L'_{f_{L^*, \P, \lambda}}(X,Y,X',Y')\left<\Phi(X,X'), \cdot\right>_\H \Phi(X,X')}{\P^2}.
	\end{align*}
\end{theorem}

For the definition of the influence function, see \citet{hampel1968, hampel1971, hampel1986}.

\begin{corollary}[Bounded Influence Function]\label{boundedInfluenceFunction}
	For all $\P \in \M_1(\X \times \Y)$, for all $(x,y,x',y') \in (\X \times \Y)^2$, and for all $\lambda \in (0, \infty)$, the influence function of $S: \M_1(\X \times \Y) \to \H$ defined by $S(\P) := f_{L^*, \P, \lambda}$ is bounded. It holds
	\begin{align*}
	\mathrm{IF}((x_0, y_0);S,\P) = -M(\P)^{-1}T(\delta_{(x_0, y_0)};\P),
	\end{align*}
	where $\delta_{(x_0, y_0)}$ denotes the Dirac distribution in the point $(x_0, y_0) \in \X \times \Y$, and $L'_{f_{L^*, \P, \lambda}}$, $T(\delta_{(x_0, y_0)};\P)$ as well as $M(\P)$ are given by \Cref{RPLGateaux}. Here $T(\delta_{(x_0, y_0)}; \P)$ simplifies to
	\begin{align*}
	T(\delta_{(x_0, y_0)};\P) &= -2\E{L'_{f_{L^*, \P, \lambda}}(X,Y,X',Y')\Phi(X,X')}{\P^2} + \\ &\qquad \E{L'_{f_{L^*, \P, \lambda}}(X,Y, x_0, y_0)\Phi(X,x_0) + L'_{f_{L^*, \P, \lambda}}(x_0, y_0,X',Y')\Phi(x_0,X')}{\P^2}
	\end{align*}
\end{corollary}

The definition of qualitative robustness was given by \citet[p.\,1890]{hampel1971} and generalized by \citet[Def.\,1, p.\,278]{cuevas1988}. We refer to \citet{cuevas1993} for the qualitative robustness of bootstrap approximations.

\begin{definition}
	A sequence of estimators $(S_n)_{n \in \N}$ is called \textbf{qualitatively robust} at a probability measure $\P$ if and only if
	\begin{align*}
	\Forall \epsilon > 0 ~ \Exists \delta > 0 ~ \Forall \Qp \in M_1(\X \times \Y): \left[d_*(\Qp,\P) < \delta \Longrightarrow d_*(\mathscr{L}_\Qp(S_n), \mathscr{L}_\P(S_n)) < \epsilon ~ \Forall n \in \N\right],
	\end{align*}
	with $\mathscr{L}_P(S_n)$ and $\mathscr{L}_Q(S_n)$ denoting the image measures $S_n \circ \P^n$ and $S_n \circ \Qp^n$ respectively, and $d_*$ being either the bounded Lipschitz metric or the Prohorov metric.
\end{definition}

Please note, that originally the Prohorov metric was used by \citet{hampel1971}. Due to the equivalence of the Prohorov metric $d_{Pro}$ and the bounded Lipschitz metric $d_{BL}$ for complete separable spaces, see e. g. \citet[Thm. 11.3.3, Cor. 11.6.5]{dudley2002}, we can also use the bounded Lipschitz metric which is easier to use in our situation, see also \citet{dudley1991}.\\

We define $\Dr_n := \frac{1}{n}\sum_{i=1}^n \delta_{(X_i,Y_i)}$ the random probability measure, and denote the distribution of the $\H$-valued RPL estimator $f_{L^*,\Dr_n, \lambda}$ by $\Ld_n(S; \P)$ for $n \in \N$. Similarly, we denote the distribution of the bootstrap approximated $\H$-valued RPL estimator $f_{L^*, \Dr, \lambda}$, when all pairs $(X_i^{(b)}, Y_i^{(b)}) \sim \Dr_n$ are independent, by $\Ld_n(S; \Dr_n)$ for $n \in \N$.

\begin{theorem}\label{continuityOperator}
	For all Borel probability measures $\P \in \M_1(\X \times \Y)$ and all $\lambda \in (0, \infty)$, we have:
	\begin{enumerate}[(i)]
		\item The RPL operator $S: \M_1(\X \times \Y) \to \H$, where $S(\P) = f_{L^*, \P, \lambda}$, is continuous with respect to the weak topology on $\M_1(\X \times \Y)$ and the norm topology on $\H$.
		\item The operator $S: \M_1(\X \times \Y) \to \Cb(\X^2)$,  where $S(\P) = f_{L^*, \P, \lambda}$, is continuous with respect to the weak topology on $\M_1(\X \times \Y)$ and the norm topology on $\Cb(\X^2)$. 
	\end{enumerate}
\end{theorem}

\begin{corollary}\label{continuityEstimator}
	For any data set $D_n \in (\X \times \Y)^n$ denote the corresponding empirical measure by $\D_n := \frac{1}{n}\sum_{i=1}^n \delta_{(x_i,y_i)}$. Then, for every $\lambda \in (0, \infty)$ and every $n \in \N$, the mapping
	$$S_n : \bigl((\X \times \Y)^n, d_{(\X \times \Y)^n}\bigr) \to (\H, d_\H), \quad S_n(D_n) = f_{L^*, \D_n, \lambda},$$
	is continuous.
\end{corollary}

\begin{theorem}[Qualitative Robustness]\label{qualitativeRobustness}
	For all $\lambda \in (0, \infty), n \in \N$ and $\Dr_n := \frac{1}{n}\sum_{i=1}^n \delta_{(X_i,Y_i)}$, we have:
	\begin{enumerate}[(i)]
		\item The sequence of RPL estimators $(S_n)_{n \in \N}$, where $S_n := f_{L^*, \Dr_n, \lambda}$, is qualitatively robust for all Borel probability measures $\P \in \M_1(\X \times \Y)$.
		\item If the metric space $\X \times \Y$ is additionally compact, then the sequence $\mathscr{L}_n(S; \Dr_n), n \in \N$, of empirical bootstrap approximations of $\mathscr{L}_n(S; \P)$ is qualitatively robust for all Borel probability measures $\P \in \M_1(\X \times \Y)$.
	\end{enumerate}
\end{theorem}

We mention that in general it is not possible to replace $\lambda$ in \Cref{qualitativeRobustness} by a null sequence $(\lambda_n)_{n \in \N}$, as there is a goal conflict between qualitative robustness and universal consistency, see \citet[Counterexample 5.2]{hable2011}.

\section{Discussion}\label{discussion}

We showed that kernel-based pairwise learning methods have good statistical robustness properties without making moment assumptions on the conditonal distribution of $Y$ given $X=x$ or boundedness assumptions on the input or output spaces. This is valid for convex Lipschitz continuous shifted loss functions and kernels which are continuous and bounded. The results can be applied in a variety of fields such as ranking, metric and online learning, we refer to for instance \citet{rejchel2012, Rej2017}, \citet{bellet2015} and \citet{ying2015}. The techniques we used are tied to those of solving nonparametric regression or classification problems with support vector machines. \\

Our work extends the results of \citet{CZ2016} to the use of prediction functions $f: \X \times \X \to \R$ with two arguments instead of restricting ourselves on the special case $f(x,x') = \tilde{f}(x) - \tilde{f}(x')$ with $\tilde{f}: \X \to \R$ being a well-investigated univariate prediction function.\\

As the present paper is on statistical robustness properties, an investigation of learning rates is beyond the scope of this paper. This also applies to the case of multivariate ranking which was already mentioned by \citet[Rem.~3, p.\,847]{CLV2008} as important problem for future research.\\

Another problem for which the theory described above could be applied to is localized learning in the same manner as for support vector machines. Optimal learning rates for localized support vector machines have been studied by \citet{steinwart2006}. The learning rates for localized classification under margin conditions have recently been improved by \citet{blaschzyk2020}. \citet{dumpert2018} have shown consistency and robustness results for the case of localized support vector machines without moment assumptions.

\appendix

\counterwithin{theorem}{subsection}

\section{Appendix}\label{appendix}

The appendix consists of one section providing definitions, theorems and lemmas which are needed for the proofs of the assertions in this paper in the second section of the appendix.

\subsection{Important definitions, theorems and lemmas}\label{appendix-1}

\begin{definition}
	Let $\X \ne \emptyset$ and $\H$ be an $\R$-Hilbert space over $\X^2$ containing functions mapping from $\X^2$ to $\R$.
	\begin{enumerate}[(a)]
		\item A function $k: \X^2 \times \X^2 \to \R$ is called a \textbf{kernel} on $\X^2$ if there exists an $\R$-Hilbert space $\H$ and a map $\Phi: \X^2 \to \H$ such that for all $(x,x'), (\tilde{x}, \tilde{x'}) \in \X^2$ we have
		\begin{align*}
		k\bigl((x,x'),(\tilde{x}, \tilde{x'})\bigr) = \left<\Phi(\tilde{x}, \tilde{x'}), \Phi(x,x')\right>_\H \stackrel{\text{in } \R}{=} \left<\Phi(x,x'), \Phi(\tilde{x}, \tilde{x'})\right>_\H
		\end{align*}
		\item A function $k: \X^2 \times \X^2 \to \R$ is called a \textbf{reproducing kernel} of $\H$ if we have $k\bigl((\cdot, \cdot), (x, x')\bigr) \in \H$ for all $(x,x') \in \X^2$ and the reproducing property
		\begin{align*}
		f(x, x') = \left<f,k\bigl((\cdot, \cdot), (x, x') \bigr)\right>_\H
		\end{align*}
		holds for all $f \in \H$.
		\item The space $\H$ is called a \textbf{reproducing kernel Hilbert space (RKHS)} over $\X^2$ if for all $(x,x') \in \X^2$ the Dirac functional $\delta_{(x, x')}: \H \to \R$ defined by
		\begin{align*}
		\delta_{(x, x')}(f) := f(x,x'), \qquad f \in \H,
		\end{align*}
		is continuous.
	\end{enumerate}
\end{definition}

It is well-known that, if $k: \X^2 \times \X^2 \to \R$ is a bounded and measurable kernel with RKHS $\H$,  $\Phi: \X^2 \to \H$ the canonical feature map and $f \in \H$ a function, then
	\begin{enumerate}[(i)]
		\item $\norm{\Phi(x,x')}_\H \leq \norm{k}_\infty, \quad \forall (x,x') \in \X^2$
		\item $\norm{f}_\infty \leq \norm{f}_\H\norm{k}_\infty$.
	\end{enumerate}

It can be shown that there is a one-to-one correspondence between reproducing kernel Hilbert spaces and kernel functions, see e.g. \citet[Theorems 4.20 and 4.21]{book}. \\

The following lemmas, see \citet[Lemmas 4.23, 4.24 and A.5.9]{book}, and theorems provide the reasoning for certain proofs.

\begin{lemma}\label{idNorm}
	Let $\X$ be a set and $k$ be a kernel on $\X^2$ with RKHS $\H$. Then $k$ is bounded if and only if every $f \in \H$ is bounded. Moreover, in this case the inclusion $\id: \H \to \ell_\infty(\X^2)$ is continuous and we have $\norm{\id: \H \to \ell_\infty(\X^2)} = \norm{k}_\infty$.
\end{lemma}

\begin{lemma}\label{measurableRKHS}
	Let $\X$ be a measurable space and $k$ be a kernel on $\X^2$ with RKHS $\H$. Then all $f \in \H$ are measurable if and only if $k\bigl((\cdot, \cdot), (x, x')\bigr): \X^2 \to \R$ is measurable for all $(
	x,x') \in \X^2$.
\end{lemma}

\begin{lemma}\label{parallelogramm}
	Let $\H$ be a Hilbert space with inner product $\left<\cdot, \cdot\right>: \H \times \H \to \R$. Then for all $f, g \in \H$, we have
	\begin{align*}
	4\left<f,g\right> &= \norm{f+g}_\H^2 - \norm{f-g}_\H^2,\\
	\norm{f+g}_\H^2+\norm{f-g}_\H^2 &= 2\norm{f}_\H^2 + 2\norm{g}_\H^2.
	\end{align*}
\end{lemma}

\begin{definition}\label{localModulusOfContinuity}
	We define the \textbf{local modulus of continuity} for the second order derivative of a loss function $L$ with respect to the last argument as
	\begin{align*}
		\omega\Arg{h}_r := \sup\bigg\{\abs{D_5D_5L(x,y,x',y',f(x,x')) - D_5D_5L(x,y,x',y',\tilde{f}(x,x'))} :\\ (x,y,x',y') \in (\X \times \Y)^2, f(x,x'), \tilde{f}(x,x') \in [-r,r], \abs{f(x,x')-\tilde{f}(x,x')} \leq h\bigg\}.
	\end{align*}
\end{definition}

The next lemma which is a consquence of \citet[Prop II.4.6]{ekeland1983} is necessary for the existence and uniqueness of a risk minimizing prediction function.

\begin{lemma}\label{ConvContMin}
	Let $E$ be a Banach space and $f: E \to \R \cup \{\infty\}$ be a convex function. If $f$ is continuous and $\lim\limits_{\norm{x}_E \to \infty} f(x) = \infty$, then $f$ has a minimizer. Moreover if $f$ is strictly convex, then $f$ has a unique minimizer in $E$.
\end{lemma}

The following proposition is a slightly modified one after Proposition 23 from \citet[p. 318]{CvMS2009} and can be proven with the same techniques.

\begin{proposition}\label{ExistenceR}
	Let $\hat{L}: (\X \times \Y)^2 \times \R \to \R$ be a measurable function which is both convex and Lipschitz continuous with respect to its fifth argument, $\P$ be a distribution on $\X \times \Y$ and $p \in [1, \infty)$. Assume that $R: \L_p(\P^2) \to \R \cup \{- \infty, \infty\}$ defined by 
	\begin{align*}
	R(g) := \int_{(\X \times \Y)^2} \hat{L}(x,y,x',y',g(x,y,x',y'))\diff \P^2(x,y,x',y')
	\end{align*}
	exists for all $f \in \L_p(\P^2)$ and define $p'$ by $\frac{1}{p} + \frac{1}{p'} = 1$. If $\abs{R(g)} < \infty$ for at least one $g \in \L_p(\P^2)$, then, for all $g \in \L_p(\P^2)$, we have
	\begin{align*}
	\partial R(g) = \{h \in \L_{p'}(\P^2): h(x,y,x',y') \in \partial \hat{L}(x,y,x',y',g(x,y,x',y')) \text{ for } \P^2\text{-almost all } (x,y,x',y')\},
	\end{align*}
	where $\partial \hat{L}(x,y,x',y',t)$ denotes the subdifferential of $\hat{L}(x,y,x',y',\cdot)$ at the point $t$.
\end{proposition}

The next statements now provide all necessities to work with subdifferentials \citet[Prop 1.11]{phelps1993} and \citet[Prop. 22]{CvMS2009}. 

\begin{proposition}\label{subdiffProp}
	Let $f: E \to \R \cup \{\infty\}$ be a convex function and $w \in E$ such that $f(w) < \infty$. If $f$ is continuous at $w$, then the subdifferential $\partial f(w)$ is a non-empty, convex and weak$^*$-compact subset of $E'$. In addition, if $c \geq 0$ and $\delta > 0$ are constants satisfying $\abs{f(v)-f(w)} \leq c\norm{v-w}_E, v \in w + \delta B_E$, then we have $\norm{w'}_E \leq c$ for all $w' \in \partial f(w)$.
\end{proposition}

\begin{lemma}\label{subdiffCalculus}
	Let $f,g: E \to \R \cup \{\infty\}$ be convex functions, $\lambda \geq 0$ and $A: F \to E$ be a bounded linear operator. We then have:
	\begin{enumerate}[(i)]
		\item For all $w \in E$ with $f(x) < \infty$, we have $\partial(\lambda f)(w) = \lambda\partial f(w)$.
		\item If there exists a $w_0 \in E$ at which $f$ is continuous, then, for all $w \in E$ satisfying both $f(w) < \infty$ and $g(w) < \infty$, we have $\partial(f+g)(w) = \partial f(w) + \partial g(w)$.
		\item If there exists a $v_0 \in F$ such that $f$ is finite and continuous at $Av_0$, then, for all $v \in F$ satisfying $f(Av) < \infty$, we have $\partial(f \circ A)(v) = A'\partial f(Av)$, where $A': E' \to F'$ denotes the adjoint operator of $A$.
		\item The function $f$ has a global minimum at $w \in E$ if and only if $0 \in \partial f(w)$.
		\item If $f$ is finite and continuous at all $w \in E$, then $\partial f$ is a monotone operator, i. e. for all $v, w \in E$ and $v' \in \partial f(v), w' \in \partial f(w)$, we have $\left<v'-w',v-w\right> \geq 0$.
	\end{enumerate}
\end{lemma}

The following theorem has been taken from \citet[Thm.~2.6]{akerkar1999} and will be used for the proof of \Cref{RPLGateaux}.

\begin{theorem}\label{partialFrechetDifferentiability}
	Let $E_1, E_2$ and $F$ be Banach spaces, $U_1 \subset E_1$ and $U_2 \subset E_2$ be open subsets and $G: U_1 \times U_2 \to F, (x_1, x_2) \mapsto G(x_1,x_2)$ be a continuous map. Then $G$ is continuously differentiable, if and only if $G$ is partially Fréchet differentiable and the partial derivatives $\frac{\partial G}{\partial x_1}$ and $\frac{\partial G}{\partial x_2}$ are continuous. In this case, the derivative of $G$ at $(x_1, x_2) \in U_1 \times U_2$ is given by
	\begin{align*}
		G'(x_1,x_2)(y_1,y_2) = \frac{\partial G}{\partial x_1}(x_1,x_2)y_1 + \frac{\partial G}{\partial x_2}(x_1,x_2)y_2, \quad (y_1,y_2) \in E_1 \times E_2.
	\end{align*}
\end{theorem}

The next theorem is a version of the classic Implicit Function Theorem by \citet[Cor.~3.4]{robinson1991}.

\begin{theorem}\label{implicit-function-thm}
	Let $E, F$ be Banach spaces and $G: E \times F \to F$ be a continuously differentiable map. Suppose that we have $(x_0, y_0) \in E \times F$ such that $G(x_0,y_0) = 0$ and $\frac{\partial G}{\partial F}(x_0,y_0)$ is invertible. Then there exist a $\delta > 0$ and a continuously differentiable map $f: x_0 + \delta B_E \to y_0 + \delta B_F$ such that for all $x \in x_0 + \delta B_E, y \in y_0 + \delta B_F$ we have $G(x,y) = 0$ if and only if $y = f(x)$. Moreover, the derivative of $f$ is given by
	\begin{align*}
		f'(x) = -\left(\frac{\partial G}{\partial F}(x,f(x))\right)^{-1}\frac{\partial G}{\partial E}(x,f(x)).
	\end{align*}
\end{theorem}

In order to show the qualitative robustness for the RPL estimator, the next theorem by \citet[Thm.~2]{cuevas1988}, which has been adapted to our notation, is useful.

\begin{theorem}\label{thm:qualitativeRobustness}
	Let $(S_n)_{n \in \N}$ be a sequence of estimators such that there exists an operator $S: \M_1(\X \times \Y) \to \H$ verifying $S_n(D_n) = S(\D_n)$ for all possible sets $\{(x_1,y_1), \dots, (x_n,y_n)\} = D_n \in (\X \times \Y)^n$ and $\D_n = n^{-1}\sum_{i=1}^n \delta_{(x_i,y_i)}$. If $S$ is continuous on $\M_1(\X \times \Y)$, then the sequence $(S_n)_{n \in \N}$ is qualitatively robust at $\P$, for all $\P \in \M_1(\X \times \Y)$.
\end{theorem}

In order to show qualitative robustness for bootstrap approximations, a result by \citet[Cor.~16.1, p.\,271]{christmann2013} is required.

\begin{theorem}\label{thm:bootstrapRobustness}
	Let $(\Omega, \A, \mu)$ be a probability space, $(\Zc, d_\Zc)$ be a compact metric space. Let $S: (\M_1(\Zc), d_{BL}) \to (\W, d_W)$ be a statistical operator with $ (\W, d_W)$ being a complete, separable metric space. Let $Z_n: (\Omega, \A, \mu) \to (\Zc, \mathcal{B}(\Zc)), n\in \N$ be independent and identically distributed random quantites and denote the image measure by $\P:= Z_n \circ \mu$. Let $S_n: (\Zc^n,d_{\Zc^n}) \to (\W, d_\W)$ be a statistic defined by $S_n(\Zc_1,\dots,\Zc_n) = S(\Dr_n)$ with $\Dr_n = \frac{1}{n}\sum_{i=1}^n \delta_{\Zc_i}$ being the corresponding (random) empirical measure. Then, if $S$ is a continuous operator, the sequence $\mathscr{L}_n(S; \Dr_n), n \in \N$, of empirical bootstrap approximations of $\mathscr{L}_n(S;\P)$ is qualitatively robust for all $\P \in \M_1(\Zc)$.
\end{theorem}

\subsection{Proofs}\label{appendix-2}

This appendix section consists of all proofs for assertions in the sections above.

\begin{Proof}{\Cref{lem:MeasurabilityLossRisk}}
	Define $e_{(x,x')} : \F \to \R, f \mapsto f(x,x')$, the evaluation map at $(x,x') \in \X^2$. Let $(f_n)_{n \in \N} \subset \F$ be a convergent sequence, such that $d(f_n, f) \to 0$
	for some $f \in \F$. Since $d$ dominates the pointwise convergence, it follows that $f_n(x, x') \to f(x, x').$
	This yields the continuity of $e_{(x,x')}$, as $e_{(x,x')}(f_n) = f_n(x,x') \to f(x,x') = e_{(x,x')}(f)$.
	
	Furthermore, the assumption $\F \subset \L_0(\X^2)$ implies that, for any $f \in \F$ the real valued map $(x,x') \mapsto f(x,x')$ defined on $\X^2$ is measurable. After applying Lemma III.14 due to \citet[p.\,70]{CV1977}, we then obtain the first assertion. The second assertion now follows from the measurability statement in Tonelli-Fubini's theorem, see \citet[p.\,148]{dudley2002}.
\end{Proof}

\begin{Proof}{\Cref{RiskConvex}}
	Let $\beta \in [0,1]$ and $f,g \in \L_0(\X^2)$, we have
	\begin{align*}
	\Risk_{L,\P}(\beta f + (1-\beta)g) &= \int_{(\X \times \Y)^2} L(x,y,x',y',\beta f(x,x') + (1-\beta)g(x,x')) \diff \P^2(x,y,x',y')\\
	&\leq \int_{(\X \times \Y)^2} \beta L(x,y,x',y',f(x,x')) + (1-\beta)L(x,y,x',y',g(x,x')) \diff \P^2(x,y,x',y')\\
	&= \beta \Risk_{L,\P}(f) + (1-\beta)\Risk_{L,\P}(g).
	\end{align*}
	In the strictly convex case, the inequality turns into a sharp one.
\end{Proof}

\begin{Proof}{\Cref{RiskConti}}
	Firstly, we show the inequality
	\begin{align*}
	\abs{\Risk_{L,\P}(f)-\Risk_{L,\P}(g)} &= \abs{\int L(x,y,x',y',f(x,x')) - L(x,y,x',y',g(x,x')) \diff \P^2(x,y,x',y')}\\
	&\leq \int \abs{L(x,y,x',y',f(x,x')) - L(x,y,x',y',g(x,x'))} \diff \P^2(x,y,x',y')\\
	&\leq \int \abs{L}_{B,1}\abs{f(x,x')-g(x,x')} \diff \P^2(x,y,x',y')\\
	&= \abs{L}_{B,1} \norm{f-g}_{\L_1(\P_X^2)}.
	\end{align*}
	Using the inequality above, the continuity of the risk functional follows immediately.
	The risk functional is well-defined as $L$ is measurable and only takes values in $\Rnp$.
\end{Proof}

\begin{Proof}{\Cref{Risk-dbar}}
	Define $L_{z,z'}(t) := L(x,y,x',y',t)$ with $z = (x,y)$ and $z' = (x',y')$ since we consider $L$ as a function of its last argument and the other four arguments are held fixed. Now let $f \in \L_\infty(\P_X^2)$ and $(f_n)_{n \in \N} \subset \L_\infty(\P_X^2)$ be a sequence with $f_n \ne 0, n \geq 1$ and $\norm{f_n}_\infty \to 0$ for $n \to \infty$. Without loss of generality, we assume that $\norm{f_n}_\infty \leq 1$ for all $n \geq 1$. For $n \geq 1$, we define
	\begin{align*}
	G_n(z,z') := \frac{L_{z,z'}(f(x,x') + f_n(x,x')) -L_{z,z'}(f(x,x'))}{f_n(x,x')} - D_5L_{z,z'}(f(x,x')),
	\end{align*}
	if $f_n(x,x') \ne 0, G_n(z,z') = 0$ else. It now follows that, for all $n \in \N$,
	\begin{align*}
	&\quad \abs{\frac{\Risk_{L,\P}(f+f_n)-\Risk_{L,\P}(f)-\Risk'_{L,\P}(f)f_n}{\norm{f_n}_\infty}}\\ 
	&\leq \int_{(\X \times \Y)^2} \abs{\frac{L_{z,z'}(f(x,x')+f_n(x,x')) - L_{z,z'}(f(x,x'))-f_n(x,x')D_5L_{z,z'}(f(x,x'))}{\norm{f_n}_\infty}} \diff \P^2(z,z')\\
	&\leq \int_{(\X \times \Y)^2} G_n(z,z')\diff \P^2(z,z').
	\end{align*}
	Since $L$ is differentiable, $G_n \to 0$ for $n \to \infty$ by definition of $f_n$. Moreover, the mean value theorem yields for $f_n(x,x') \ne 0$, that there exists a function $g_n: (\X \times \Y)^2 \to \R$ with $\abs{g_n(z,z')} \in [0, f_n(x,x')]$ and
	\begin{align*}
	\frac{L_{z,z'}(f(x,x'))+f_n(x,x'))-L_{z,z'}(f(x,x'))}{f_n(x,x')} = D_5L_{z,z'}\bigl(f(x,x')+g_n(z,z')\bigr).
	\end{align*}
	Since $\abs{D_5L}$ is a $\P$-integrable Nemitski loss, it follows for all $(x, y, x', y') \in (\X \times \Y)^2$ that 
	\begin{align*}
	\abs{D_5L(x,y,x',y',t)} \leq b(x,y,x',y') + h(\abs{t}),
	\end{align*}
	with $b \in \L_1(\P^2)$ and $h: \Rnp \to \Rnp$ an increasing function. Combining these two considerations, we get
	\begin{align*}
	\abs{\frac{L_{z,z'}(f(x,x'))+f_n(x,x'))-L_{z,z'}(f(x,x'))}{f_n(x,x')}} &\leq b(x,y,x',y')+h\Arg{\abs{f(x,x')+g_n(x,y,x',y')}}\\
	&\leq b(x,y,x',y') + h\Arg{\norm{f}_\infty + 1},
	\end{align*}
	for all $n \geq 1$ with $f_n(x,x') \ne 0$. It follows for all $(x,y,x',y') \in (\X \times \Y)^2$ that
	\begin{align*}
	0 \leq G_n(x,y,x',y') \leq 2b(x,y,x',y')+2h\Arg{\norm{f}_\infty+1}.
	\end{align*}
	The assertion now follows from Lebesgue's theorem of dominated convergence.
\end{Proof}

\begin{Proof}{\Cref{shifted-convex}}
	Follows immediately from the definition of a convex or Lipschitz continuous pairwise loss function, respectively.
\end{Proof}

\begin{Proof}{\Cref{Inequalities}}
	\begin{enumerate}[(i)]
		\item We immediately obtain \begin{align*}
		\inf\limits_{t \in \R} L^*(x, y, x',y',t) \leq L^*(x,y,x',y',0) = L(x,y,x',y',0) - L(x,y,x',y',0) = 0.
		\end{align*}
		\item For all $f \in \H$, we have
		\begin{align*}
		\abs{\Risk_{L^*,\P}(f)} &= \abs{\E{L^*(X,Y,X',Y',f(X,X'))}{\P^2}}\\
		 &\leq \E{\abs{L(X,Y,X',Y',f(X,X')) - L(X,Y,X',Y',0)}}{\P^2}\\
		&\leq \abs{L}_1\E{\abs{f(X,X')}}{\P^2},
		\end{align*}
		which proves (2.2). The inequality (2.3) follows from \Cref{regRisk} and the calculations given above.
		\item As $0 \in \H$, we obtain
		\begin{align*}
		\inf\limits_{f \in \H} \Risk^\text{reg}_{L^*,\P, \lambda}(f) \leq  \Risk^\text{reg}_{L^*,\P, \lambda}(0) = 0 = \Risk_{L^*, \P}(0),
		\end{align*}
		this yields $\inf\limits_{f \in \H} \Risk_{L^*,\P}(f) \leq 0.$
		\item Due to (iii) $\Risk^\text{reg}_{L^*,\P, \lambda}(f_{L^*,\P,\lambda}) \leq 0$. As $L$ is a non-negative function, we obtain
		\begin{align*}
		\lambda\norm{f_{L^*,\P,\lambda}}_\H^2 &\leq -\Risk_{L^*,\P}(f_{L^*,\P,\lambda})\\
		&= \E{L(X,Y,X',Y',0) - L(X,Y,X',Y',f_{L^*,\P,\lambda}(X,X'))}{\P^2}\\
		&\leq \E{L(X,Y,X',Y',0)}{\P^2} = \Risk_{L,\P}(0)
		\end{align*}
		and thus (2.4) follows. To prove (2.5), we consider
		\begin{align*}
		0 &\leq -\Risk^\text{reg}_{L^*,\P, \lambda}(f_{L^*,\P,\lambda})\\
		&= \E{L(X,Y,X',Y',0) - L(X,Y,X',Y',f_{L,\P,\lambda}(X,X'))}{\P^2} - \lambda\norm{f_{L^*,\P,\lambda}}_\H^2\\
		&\hspace*{-0.3em}\stackrel{L \geq 0}{\leq} \E{L(X,Y,X',Y',0)}{\P^2} = \Risk_{L,\P}(0).
		\end{align*}
		Furthermore, we obtain
		\begin{align*}
		-\abs{L}_1\E{\abs{f_{L^*,\P,\lambda}(X,X')}}{\P^2} + \lambda\norm{f_{L^*,\P,\lambda}}_\H^2 &\leq \Risk^\text{reg}_{L^*,\P, \lambda}(f_{L^*,\P,\lambda}) \leq \Risk^\text{reg}_{L^*,\P, \lambda}(0) = 0,
		\end{align*}
		which yields (2.6). Using (2.6) and the reproducing property, we get for $f_{L^*,\P,\lambda} \ne 0$ that
		\begin{align*}
		\norm{f_{L^*,\P,\lambda}}_\infty \leq \norm{k}_\infty\norm{f_{L^*,\P,\lambda}}_\H &\leq \norm{k}_\infty \sqrt{\lambda^{-1} \abs{L}_1\E{\abs{f_{L^*,\P,\lambda}(X,X')}}{\P^2}}\\
		&\leq \norm{k}_\infty \sqrt{\lambda^{-1}\abs{L}_1\norm{f_{L^*,\P,\lambda}}_\infty},
		\end{align*}
		which is finite as $k$ is a bounded kernel. Hence $\norm{f_{L^*,\P,\lambda}}_\infty \leq \norm{k}_\infty^2\lambda^{-1}\abs{L}_1$. The case $f_{L^*,\P,\lambda} = 0$ is trivial. The inequality (2.8) now follows immediately, as 
		\begin{align*}
		\Risk_{L^*,\P}(f_{L^*,\P,\lambda}) &= \E{L(X,Y,X',Y',f_{L^*,\P,\lambda}(X,X')) - L(X,Y,X',Y',0)}{\P^2}\\
		&\leq \E{\abs{L}_1\abs{f_{L^*,\P,\lambda}(X,X')-0}}{\P^2}\\
		&= \abs{L}_1 \E{\abs{f(X,X')}}{\P^2}\\
		&\leq \abs{L}_1 \norm{f_{L^*,\P,\lambda}}_\infty\\
		&\leq \lambda^{-1}\abs{L}_1^2\norm{k}_\infty^2.
		\end{align*}
		\item 
		We have, for all $(x,y,x,y',t) \in (\X \times \Y)^2 \times \R$,
		\begin{align*}
		\hspace*{-1cm}
		D_5 L^*(x,y,x',y',t) &= \lim_{\substack{h \to 0 \\ h \ne 0}} \frac{L^*(x,y,x',y',t+h)-L^*(x,y,x',y',t)}{h}\\
		&= \lim_{\substack{h \to 0 \\ h \ne 0}} \frac{L(x,y,x',y',t+h) - L(x,y,x',y',0) -L(x,y,x',y',t) + L(x,y,x',y',0)}{h}\\
		&= \lim_{\substack{h \to 0 \\ h \ne 0}} \frac{L(x,y,x',y',t+h)-L(x,y,x',y',t)}{h}\\
		&= D_5 L(x,y,x',y',t). \qedhere
		\end{align*}
		
	\end{enumerate}
\end{Proof}

\begin{Proof}{\Cref{LipschitzRisk}}
	Using (2.2) from \Cref{Inequalities}, we have \begin{align*}
	\abs{\Risk_{L^*,P}(f)} \leq \abs{L}_1\E{\abs{f(X,X')}}{\P^2_X} < \infty,
	\end{align*}
	for any $f \in \L_1(\P_X^2)$. Inequality (2.3) yields that 
	\begin{align*}
	\Risk^\text{reg}_{L^*,\P, \lambda}(f) &\geq -\abs{L}_1\E{\abs{f(X,X')}}{\P^2_X} + \lambda\norm{f}_\H^2 > -\infty. \qedhere
	\end{align*}
\end{Proof}

\begin{proof}[Proof of \Cref{thm:uniquenessMinimizer}]
	Let us assume that the mapping $f \mapsto \lambda\norm{f}_\H^2 + \Risk_{L^*,\P}$ has two minimizers $f_1, f_2 \in \H$ with $f_1 \ne f_2$. 
	\begin{enumerate}[(i)]
		\item By the parallelogram identity, we then find 
		\begin{align*}
		\norm{\frac{1}{2}(f_1+f_2)}_\H^2 < \frac{1}{2}\left(\norm{f_1}_\H^2 + \norm{f_2}_\H^2\right).
		\end{align*}
		As $L$ is convex, $L^*$ and $\Risk_{L^*,\P}$ are also convex due to \Cref{shifted-convex} and \Cref{RiskConvex}.
		The convexity of the map $f \mapsto \Risk_{L^*,\P}(f)$ and
		\begin{align*}
		\lambda\norm{f_1}_\H^2 + \Risk_{L^*,\P}(f_1) = \lambda\norm{f_2}_\H^2 + \Risk_{L^*,\P}(f_2)
		\end{align*}
		yield for $f^* := \frac{1}{2}(f_1+f_2)$ that
		\begin{align*}
		\lambda\norm{f^*}_\H^2+\Risk_{L^*,\P}(f^*) &< \frac{\lambda}{2}\left(\norm{f_1}_\H^2+\norm{f_2}_\H^2\right) + \frac{1}{2}\Risk_{L^*,\P}(f_1) + \frac{1}{2}\Risk_{L^*,\P}(f_2)\\
		&< 	\lambda\norm{f_1}_\H^2 + \Risk_{L^*,\P}(f_1),
		\end{align*}
		i.e. $f_1$ is not a minimizer of $f \mapsto \lambda\norm{f}_\H^2 + \Risk_{L^*,\P}(f)$. Consequently, the assumption that there are two minimizers is false by contradiction.
		\item This condition implies $\abs{\Risk_{L^*,\P}} < \infty$ due to \Cref{LipschitzRisk} and the assertion follows from (i). \qedhere
	\end{enumerate}
\end{proof}

\begin{Proof}{\Cref{thm:existenceMinimizer}}
	Since the kernel $k$ is measurable, its RKHS $\H$ consists of measurable functions. Moreover, $k$ is bounded and thus $\id: \H \to \L_\infty(\P_X^2)$ is continuous. Additionally, $L$ is non-negative and hence $-\infty < L^*(x,y,x',y',t) < \infty$ for all $(x,y,x',x',t) \in (\X \times \Y)^2 \times \R$. Thus $L^*$ is continuous by the convexity of $L^*$ with respect to the fifth argument. Therefore, \Cref{RiskConti} yields that $\Risk_{L^*,\P}: \L_\infty(\P_X^2) \to \R$ is continuous and hence $\Risk_{L^*,\P}: \H \to \R$ is continuous, because $\H \subset \L_\infty(\P_X^2)$. Furthermore, \Cref{RiskConvex} provides the convexity of this mapping. It follows that $f \mapsto \lambda\norm{f}_\H^2 + \Risk_{L^*,\P}(f)$ is convex, because $f \mapsto \lambda\norm{f}_\H^2$ is convex. \Cref{ConvContMin} shows that if $\Risk^\text{reg}_{L^*,\P, \lambda}(f)$ is convex and continuous and additionally $\Risk^\text{reg}_{L^*,\P, \lambda}(f) \to \infty$ for $\norm{f}_\H \to \infty$, then $\Risk^\text{reg}_{L^*,\P, \lambda}(\cdot)$ has a minimizer. Therefore it is only left to show that this limit is infinite. We have
	\begin{align*}
	\Risk^\text{reg}_{L^*,\P, \lambda}(f) &\stackrel{(2.1)}{\geq} -\abs{L}_1\E{\abs{f(X,X')}}{\P_X^2} + \lambda\norm{f}_\H^2\\
	&\geq -\abs{L}_1\norm{f}_\infty + \lambda\norm{f}_\H^2\\
	&\overset{\mathclap{\text{rep. property}}}{\geq} \hspace*{0.5em} -\abs{L}_1\norm{k}_\infty\norm{f}_\H + \lambda\norm{f}_\H^2 \to \infty,
	\end{align*}
	for $\norm{f}_\H  \to \infty$, as $\abs{L}_1\norm{k}_\infty \in \Rnp$ and $\lambda > 0$.
\end{Proof}

We need the following auxiliary lemma in order to prove the existence of minimizers in the non-convex case.

\begin{lemma}\label{lem:subsequence-HilbertSpace}
	Let $r \in (0, \infty)$. If $f_0 \in \H$ and if the sequence $(f_\ell)_{\ell \in \N} \subset B_r(f_0) := \{f \in \H: \norm{f-f_0}_\H \leq r\}$, then there exists a subsequence $(f_{\eta(\ell)})_{\ell \in \N} \subset (f_\ell)_{\ell \in \N} \subset B_r(f_0)$ with $\eta: \N \to \N$ increasing and $f^* \in B_r(f_0)$ such that 
	$$\norm{f^*}_\H \leq \liminf_{\ell \to \infty} \norm{f_{\eta(\ell)}}_\H$$ and
	$$\lim_{\ell \to \infty} f_{\eta(\ell)}(x, x') = f^*(x, x'), \quad \Forall (x, x') \in \X^2.$$
\end{lemma}

\begin{Proof}{\Cref{lem:subsequence-HilbertSpace}}
	The closed ball $B_r(f_0) \subset \H$ is weakly compact and hence there exists a subsequence $(f_{\eta(\ell)})_{\ell \in \N} \subset B_r(f_0)$ weakly converging to some $f^* \in B_r(f_0)$, i.e.
	\begin{align*}
	\lim_{\ell \to \infty} \left<f_{\eta(\ell)}, f\right>_\H = \left<f^*,f\right>_\H, \quad \Forall f \in \H.
	\end{align*}
	Let $f = f^*$, then by using the Cauchy-Schwarz inequality, it follows
	\begin{align*}
	\norm{f^*}_\H^2 = \left<f^*, f^*\right>_\H = \lim_{\ell \to \infty} \left<f_{\eta(\ell)}, f^*\right>_\H \leq \liminf_{\ell \to \infty} \norm{f_{\eta(\ell)}}_\H\norm{f^*}_\H,
	\end{align*}
	which implies $\norm{f^*}_\H \leq \liminf_{\ell \to \infty} \norm{f_{\eta(\ell)}}_\H.$
	Let $f = \Phi(x, x'), (x, x') \in \X^2$, then the reproducing property yields the remaining assertion
	\begin{align*}
	\lim_{\ell \to \infty} f_{\eta(\ell)}(x,x') = \lim_{\ell \to \infty} \left<f_{\eta(\ell)}, \Phi(x, x')\right>_\H = \left<f^*, \Phi(x,x')\right>_\H &= f^*(x,x'). \qedhere
	\end{align*}
\end{Proof}

\begin{Proof}{\Cref{thm:existenceMinimizerNonConvex}}
	For every $\ell \in \N$, set $f_\ell \in \H$ such that
	\begin{align}
	\Risk^\text{reg}_{L, \P, \lambda}(f_\ell) = \Risk_{L,\P}(f_\ell) + \lambda\norm{f_\ell}^2_\H &\leq \inf_{f \in \H} \Risk_{L,\P}(f) + \lambda\norm{f}_\H^2 + \frac{1}{\ell}.
	\end{align}
	Taking $f = f_0$, we conclude that
	\begin{align*}
	\lambda\norm{f_\ell}_\H^2 \leq \Risk_{L,\P}(f_0) + \lambda\norm{f_0}^2_\H + 1
	\end{align*}
	and thus $f_\ell \in B_r(0) = \{f \in \H: \norm{f}_\H \leq r\}$ with $r := \frac{\Risk_{L,\P}(f_0) + \lambda\norm{f_0}^2_\H + 1}{\lambda}$. Application of \Cref{lem:subsequence-HilbertSpace} yields that there exists a subsequence $(f_{\eta(\ell)})_{\ell \in \N} \subset B_r(0)$ and some $f^* \in B_r(0)$ such that $\norm{f^*}_\H \leq \liminf_{\ell \to \infty} \norm{f_{\eta(\ell)}}_\H$ and $f_{\eta(\ell)}(x, x') \to f^*(x, x')$ for all $(x, x') \in \X^2$. By the Lipschitz continuity of $L$, it follows
	\begin{align*}
	\abs{L(x,y,x',y',f_{\eta(\ell)})-L(x,y,x',y',f_0(x,x'))} &\leq \abs{L}_1\abs{f_{\eta(\ell)}(x,x')-f_0(x,x')}\\
	&\leq \abs{L}_1 \cdot \Arg{\abs{f_{\eta(\ell)}(x,x')} + \abs{f_0(x,x')}}\\
	&\leq \abs{L}_1 \Arg{\norm{f_{\eta(\ell)}}_\infty + \norm{f_0}_\infty}\\
	&\leq \abs{L}_1 \Arg{\norm{f_{\eta(\ell)}}_\H + \norm{f_0}_\H} \cdot \norm{k}_\infty\\
	&\leq \abs{L}_1 \Arg{r + r} \cdot \norm{k}_\infty\\
	&= 2r\abs{L}_1\norm{k}_\infty.
	\end{align*}  
	Therefore
	\begin{align*}
	L(x,y,x',y',f_{\eta(\ell)}(x,x')) &\leq L(x,y,x',y',f_0(x,x')) + 2r\abs{L}_1\norm{k}_\infty < \infty\\
	L(x,y,x',y',f_{\eta(\ell)}(x,x')) &\geq -L(x,y,x',y',f_0(x,x')) - 2r\abs{L}_1\norm{k}_\infty > -\infty
	\end{align*}
	with the upper and lower bound being $\P^2$-integrable. Since $f_{\eta(\ell)} \to f^*$ pointwise for every $(x,x') \in \X^2$, we have by the continuity of $L$
	\begin{align*}
	\lim_{\ell \to \infty} L(x,y,x',y',f_{\eta(\ell)}(x,x')) = L(x,y,x',y',f^*(x,x')).
	\end{align*}
	Lebesgue's theorem of dominated convergence yields $\lim_{\ell \to \infty} \Risk_{L,\P}(f_{\eta(\ell)}) = \Risk_{L,\P}(f^*)$.
	Taking the limit inferior on both sides of inequality (A.1) gives the result
	\begin{align*}
	\Risk^\text{reg}_{L,\P, \lambda}(f^*) \leq \inf_{f \in \H} \Risk_{L,\P}(f)+ \lambda\norm{f}_\H^2,
	\end{align*}
	which means that $f^*$ is a minimizer for the regularized risk.
\end{Proof}

\begin{Proof}{\Cref{representerTheorem}}
	The existence and uniqueness of $f_{L^*,\P,\lambda}$ follow from \Cref{thm:uniquenessMinimizer} and \Cref{thm:existenceMinimizer}. As $k$ is bounded, \Cref{Inequalities}(iv) is applicable and inequalities (6) and (7) yield 
	\begin{align*}
	\norm{f_{L^*, \P, \lambda}}_\infty &\leq \lambda^{-1}\abs{L}_1\norm{k}_\infty^2 < \infty\\
	\text{and} \quad \abs{\Risk_{L^*,\P}(f_{L^*,\P,\lambda})} &\leq \lambda^{-1}\abs{L}^2_1\norm{k}_\infty^2 < \infty.
	\end{align*}
	Furthermore, due to \Cref{shifted-convex}(ii) $L^*$ is a Lipschitz continuous pairwise loss function, because $L$ is given as such. Define $R: \L_1(\P^2) \to \R$ by
	\begin{align*}
	R(g) := \int_{(\X \times \Y)^2} L^*(x,y,x',y',g(x,y,x',y')) \diff \P^2(x,y,x',y').
	\end{align*}
	The operator $R$ is well-defined, because due to the Lipschitz continuity of $L^*$ with respect to its fifth argument, we obtain
	\begin{align*}
	\abs{R(g)} \leq \abs{L}_1\int_{(\X \times \Y)^2} \abs{g(x,y,x',y')}\diff \P^2(x,y,x',y') < \infty,
	\end{align*}
	since $g \in \L_1(\P^2)$. The continuity of $R$ can be shown as follows. Fix $\delta > 0$ and let $f_1, f_2 \in \L_1(\P^2)$ with $\norm{f_1-f_2}_{\L_1(\P^2)} < \delta$. The Lipschitz continuity of $L^*$ yields
	\begin{align*}
	\abs{R(f_1) - R(f_2)} &\leq \int_{(\X \times \Y)^2} \abs{L^*(x,y,x',y',f_1(x,x')) - L^*(x,y,x',y',f_2(x,x'))} \diff \P^2(x,y,x',y')\\
	&\leq \abs{L}_1 \int_{(\X \times \Y)^2}\abs{f_1(x,y,x',y')-f_2(x,y,x',y')} \diff \P^2(x,y,x',y')\\
	&< \delta\abs{L}_1,
	\end{align*}
	and so the continuity of $R$. We can now apply \Cref{ExistenceR} with $p=1$, because $R(f)$ exists and is well-defined for all $g \in \L_1(\P^2)$. The subdifferential of $R$ can thus be computed by 
	\begin{align*}
	\partial R(g) = \{h \in \L_\infty(\P^2): h(x,y,x',y') \in \partial L^*(x,y,x',y',g(x,y,x',y')) \text{ for } \P^2\text{-almost all } (x,y,x',y')\}.
	\end{align*}
	Now, we infer from \Cref{idNorm} that the inclusion map $I: \H \to \L_1(\P_X^2)$ defined by
	\begin{align*}
	(If)(x,y,x',y') := f(x,x')
	\end{align*}
	is a bounded linear operator. Furthermore, $S: \H \to \R,\, S(g) := \left<f,g\right>_\H$ is a bounded linear operator and it follows that $S(\E{g}{\P^2}) = \E{S(g)}{\P^2}$ for bounded linear operators and Bochner integrals, see e.g. \citet[Thm.~3.10.16]{denkowski2014}. Moreover, for all $h \in \L_\infty(\P^2)$ and all $f \in \H$, the reproducing property yields with $\Phi: \X^2 \to \H: \Phi(x,x') := k\bigl((\cdot, \cdot), (x,x')\bigr)$ the canonical feature map:
	\begin{align*}
	\left<h, If\right>_{\L_\infty(\P^2), \L_1(\P^2)} &:= \E{hIf}{\P^2}\\
	&= \int_{(\X \times \Y)^2} h(x,y,x',y')(If)(x,y,x',y')\diff \P^2(x,y,x',y')\\
	&=\int_{(\X \times \Y)^2} h(x,y,x',y')f(x,x')\diff \P^2(x,y,x',y')\\
	&= \int_{(\X \times \Y)^2} h(x,y,x',y')\left<f,\Phi(x,x')\right>_\H \diff \P^2(x,y,x',y')\\
	&= \E{h\left<f,\Phi\right>_\H}{\P^2}= \left<f, \E{h\Phi}{\P^2}\right>_\H = \left<\iota\E{h\Phi}{\P^2}, f\right>_{\H', \H}\,,
	\end{align*}
	with $\iota: \H \to \H'$ the Fréchet-Riesz isomorphism, see e.g. \citet[Thm.~V.3.6]{werner2011}. Thus the adjoint operator $I'$ of $I$ is given by 
	\begin{align*}
	I'h = \iota\E{h\Phi}{\P^2}, \quad h \in \L_\infty(\P^2).
	\end{align*}
	Moreover, the $L^*$-risk functional $\Risk_{L^*,\P}: \H \to \R$ satisfies
	\begin{align*}
	\Risk_{L^*,\P} = R \circ I
	\end{align*}
	and hence the chain rule for subdifferentials, \Cref{subdiffCalculus}(iii), see also \citet[Thm.~5.3.33]{denkowski2014}, yields
	\begin{align*}
	\partial\Risk_{L^*,\P}(f) = \partial(R \circ I)(f) = I'\partial R(If),
	\end{align*}
	for all $f \in \H$. Applying the formula for $\partial R(f)$ thus yields, for all $f \in \H$
	\begin{align*}
	\partial\Risk_{L^*,\P}(f) = \{\iota\E{h\Phi}{\P^2}: h \in \L_\infty(\P^2) \text{ with } h(x,y,x',y') \in \partial L^*(x,y,x',y',f(x,x')) ~ \P^2\text{-a.s.}\}.
	\end{align*} 
	 In addition, $f \mapsto \norm{f}^2_\H$ is Fréchet-differentiable and its derivative at $f$ is $2\iota f$ for all $f \in \H$. By picking suitable representations of $h \in \L_\infty(\P^2)$, \Cref{subdiffCalculus} thus gives for all $f \in \H$
	\begin{align*}
	\partial\Risk^\text{reg}_{L^*,\P, \lambda}(f) = 2\lambda \iota f + \{&\iota\E{h\Phi}{\P^2}: h \in \L_\infty(\P^2) \text{ with } h(x,y,x',y') \in \partial L^*(x,y,x',y',f(x,x'))\\
	&\Forall(x,y,x',y') \in (\X \times \Y)^2\}
	\end{align*}
	for all $f \in \H$. Now recall that $\Risk^\text{reg}_{L^*,\P, \lambda}(\cdot)$ has a minimum at $f_{L^*, \P, \lambda} \in \H$ and therefore we have $0 \in \partial \Risk^\text{reg}_{L^*,\P, \lambda}(f_{L^*, \P, \lambda})$ by \Cref{subdiffCalculus}(iv). This together with the injectivity of $\iota$ yields the assertions (i) and (ii). Let us now show that (iii) is valid. Since $k$ is a bounded kernel, we have by the second part of \Cref{Inequalities}(iv)
	\begin{align*}
	\norm{f_{L^*,\P,\lambda}}_\infty \leq  \lambda^{-1}\abs{L}_1\norm{k}_\infty^2 =: B_\lambda < \infty.
	\end{align*}
	Now (i) and \Cref{subdiffProp} with $\delta := 1$ yield, for all $(x,y,x',y') \in (\X \times \Y)^2$,
	\begin{align*}
	\abs{h(x,y,x',y')} \leq \sup_{(x,y,x',y') \in (\X \times \Y)^2} \abs{\partial L^*(x,y,x',y',f_{L^*,\P,\lambda}(x,x'))} \leq \abs{L}_1.
	\end{align*}
	Hence $h \in \L_\infty(\P_X^2) $ and the assertion (iii) follows. \\
	
	To prove (iv), we use (i) and the definition of the subdifferential to obtain, for all $(x,y,x',y') \in (\X \times \Y)^2$,
	\begin{align*}
	&\hspace*{1.3em} h_\P(x,y,x',y')(f_{L^*,\Qp,\lambda}(x,x') - f_{L^*,\P,\lambda}(x,x'))\\ 
	&\leq L^*(x,y,x',y'f_{L^*,\Qp,\lambda}(x,x')) - L^*(x,y,x',y',f_{L^*,\P,\lambda}(x,x')).
	\end{align*}
	By integrating with respect to $\Qp$, we hence obtain
	\begin{align*}
	\left<f_{L^*,\Qp,\lambda} - f_{L^*,\P,\lambda}, \E{h_\P\Phi}{\Qp^2}\right>_\H &\leq \Risk_{L^*,\Qp}(f_{L^*,\Qp,\lambda}) - \Risk_{L^*,\Qp}(f_{L^*,\P,\lambda}).
	\end{align*}
	Moreover, an easy calculation yields
	\begin{align*}
	&\hspace*{1.2em}\left<f_{L^*,\Qp,\lambda}-f_{L^*,\P,\lambda}, \E{h\Phi}{\Qp^2} + 2\lambda f_{L^*,\P,\lambda}\right>_\H + \lambda\norm{f_{L^*,\P,\lambda}-f_{L^*,\Qp,\lambda}}_\H^2\\ 
	&\leq \Risk_{L^*,\Qp, \lambda}^\text{reg}(f_{L^*,\Qp,\lambda}) - \Risk_{L^*,\Qp, \lambda}^\text{reg}(f_{L^*,\P,\lambda}) \leq 0,
	\end{align*}
	and consequently using the representation $f_{L^*,\P,\lambda} = -(2\lambda)^{-1}\E{h\Phi}{\P^2}$,
	it follows after using the Cauchy-Schwarz inequality that
	\begin{align*}
	\lambda\norm{f_{L^*,\P,\lambda} - f_{L^*,\Qp,\lambda}}_\H^2 &\leq \left<f_{L^*,\P,\lambda}-f_{L^*,\Qp,\lambda}, \E{h_\P\Phi}{\Qp^2}-\E{h\Phi}{\P^2}\right>_\H\\
	&\leq \norm{f_{L^*,\P,\lambda}-f_{L^*,\Qp,\lambda}}_\H \norm{\E{h_\P\Phi}{\Qp^2} - \E{h\Phi}{\P^2}}_\H.
	\end{align*}
	This yields the last assertion.
\end{Proof}

In order to prove the risk consistency, a formulation of Hoeffding's inequality for Hilbert spaces is required. The original inequality can be found in \citet[p.\,6]{hoeffding1963}, see e.g. \citet[p.\,217]{book} for the Hilbert space version.

\begin{theorem}[Hoeffding's inequality in Hilbert spaces]\label{HoeffdingIneq}
	Let $(\Omega, \A, \P)$ be a probability space, $\H$ be a separable Hilbert space and $B > 0$. Furthermore, let $\xi_1, \dots, \xi_n: \Omega \to \H$ be independent random variables satisfying $\norm{\xi_i}_\infty \leq B$ for all $1 \leq i \leq n$. Then, for all $\tau > 0$, we have
	\begin{align*}
	\P\Arg{\norm{n^{-1}\sum_{i=1}^{n}\Arg{\xi_i-\E{\xi_i}{P}}}_\H \geq B\sqrt{\frac{2\tau}{n}} + B\sqrt{\frac{1}{n}}+\frac{4B\tau}{3n}} \leq e^{-\tau}.
	\end{align*}
\end{theorem}

\begin{Proof}{\Cref{riskConsistency}}
	Without loss of generality, let $\norm{k}_\infty = 1$. This implies $\norm{f}_\infty \leq \norm{f}_\H$ for all $f \in \H$. Let $L, L^* \ne 0$. The Lipschitz continuity of the pairwise loss functions $L$ and thus $L^*$, \Cref{RiskConti}, and the following remark yield for all $g \in \H$ and for all $n \in \N$,
	\begin{align*}
	\abs{\Risk_{L^*,\P}(f_{L^*,\P,\lambda_n}) - \Risk_{L^*,\P}(g)} \leq \abs{L}_1\norm{f_{L^*,\P,\lambda_n} - g}_\H.
	\end{align*}
	For $n \in \N$ and $\lambda_n > 0$, let $h_n := h_{L^*, \P, n}: (\X \times \Y)^2 \to \R$ be the function obtained by the Representer Theorem \ref{representerTheorem}. Let $\Phi: \X^2 \to \H$ be the canonical feature map of $k$. The Representer Theorem yields for all $\Qp \in \M_1(\X \times \Y)$, that
	\begin{align*}
	\norm{f_{L^*, \P, \lambda_n} - f_{L^*, \Qp, \lambda_n}}_\H \leq \lambda_n^{-1}\norm{\E{h_n\Phi}{\P^2} - \E{h_n\Phi}{\Qp^2}}_\H.
	\end{align*}
	Recall that the function $h_n$ may depend on $\P$, but it is independent of $\Qp$.
	Moreover, let $\epsilon \in (0,1)$ and $D$ be a training set of $n$ data points with corresponding empirical distribution $\D$ such that
	\begin{align}\label{UngleichungErw}
	\norm{\E{h_n\Phi}{\P^2} - \E{h_n\Phi}{\D^2}} \leq \frac{\lambda_n\epsilon}{\abs{L}_1}.
	\end{align}
	It follows that 
	\begin{align*}
	\norm{f_{L^*, \P, \lambda_n} - f_{L^*, \D, \lambda_n}}_\H \leq \frac{\epsilon}{\abs{L}_1}
	\end{align*}
	and hence
	\begin{align*}
	\abs{\Risk_{L^*,\P}(f_{L^*,\P,\lambda_n}) - \Risk_{L^*, \P}(f_{L^*,\D,\lambda_n})} \leq \abs{L}_1 \norm{f_{L^*,\P,\lambda_n} - f_{L^*,\D,\lambda_n}}_\H \leq \epsilon.
	\end{align*}
	We will now determine the probability of the training set $D$ to satisfy \Cref{UngleichungErw}. The assumption $\lambda_n^2n \to \infty$ implies that $\lambda_n\epsilon \geq n^{-1/2}$ for sufficiently large $n \in \N$. The third statement of the Representer Theorem \ref{representerTheorem} shows that $\norm{h_n}_\H \leq \abs{L}_1$ and our assumption $\norm{k}_\infty = 1$ yields $\norm{h_n\Phi}_\infty \leq \abs{L}_1$. 
	Set $\zeta_n := \abs{L}_1^{-1}\lambda_n \epsilon$. The fact that $\lambda_n \epsilon \geq \frac{1}{\sqrt{n}}$ for sufficiently large $n$ implies that 
	\begin{align*}
		\frac{\zeta_n}{3} = \frac{\lambda_n\epsilon}{3\abs{L}_1} > \frac{1}{\sqrt{n}},
	\end{align*}
	for sufficiently large $n$. For 
	\begin{align*}
		\tau = \frac{3}{8} \cdot \frac{\abs{L}_1^{-2}\epsilon^2\lambda_n^2n}{\abs{L}_1^{-1}\epsilon\lambda_n +3} > 0,
	\end{align*} the following inequality holds using the calculation above
	\begin{align*}
	\frac{1}{\sqrt{n}}(\sqrt{2\tau} + 1) + \frac{4\tau}{3n} &= \frac{1}{\sqrt{n}}\Arg{\sqrt{\frac{3}{4} \cdot \frac{\zeta_n^2n}{\zeta_n + 3}} + 1} + \frac{\zeta_n^2}{2(\zeta_n+3)}\\
	&= \frac{\sqrt{3\zeta_n^2}}{2\sqrt{\zeta_n+3}} + \frac{1}{\sqrt{n}} + \frac{\zeta_n}{2} \cdot \frac{\zeta_n}{\zeta_n+3}\\
	&= \frac{\zeta_n}{2} \cdot \underbrace{\frac{\sqrt{3}}{\sqrt{\zeta_n+3}}}_{< 1} + \frac{1}{\sqrt{n}} + \frac{\zeta_n}{2} \cdot \underbrace{\frac{\zeta_n}{\zeta_n+3}}_{< 1/3}\\
	&< \frac{\zeta_n}{2} + \frac{1}{\sqrt{n}} + \frac{\zeta_n}{6}\\
	&< \zeta_n.
	\end{align*}
	
	An application of Hoeffding's inequality, i.e. \Cref{HoeffdingIneq}, for the case $B = 1$ yields
	\begin{align*}
		&\quad\, \P\Arg{\norm{\E{h_n\Phi}{\D^2} - \E{h_n\Phi}{\P^2}}_\H \leq \frac{\lambda_n\epsilon}{\abs{L}_1}}\\	
		&= \P\Arg{\norm{n^{-1}\sum_{i=1}^n \Arg{h_n(X_i,Y_i,X'_i,Y'_i)\Phi(X_i,X'_i) - \E{h_n\Phi}{\P^2}}}_\H \leq \frac{\lambda_n\epsilon}{\abs{L}_1}}\\
		&\geq \P\Arg{\norm{n^{-1}\sum_{i=1}^n \Arg{h_n(X_i,Y_i,X'_i,Y'_i)\Phi(X_i,X'_i) - \E{h_n\Phi}{\P^2}}}_\H \leq n^{-1/2}(\sqrt{2\tau} + 1) + \frac{4\tau}{3n}}\\
		&\geq 1-\exp\Arg{-\tau} = 1-\exp\Arg{-\frac{3}{8} \cdot \frac{\abs{L}_1^{-2}\epsilon^2\lambda_n^2n}{\abs{L}_1^{-1}\epsilon\lambda_n +3}} = 1-\exp\Arg{-\frac{3}{8} \cdot \frac{\epsilon^2\lambda_n^2n}{(\epsilon\lambda_n + 3\abs{L}_1)\abs{L}_1}},
	\end{align*}
	for sufficiently large $n \in \N$.
	Using the regularity assumptions, it follows that the probability converges to $1$, if $\abs{D} = n \to \infty$. This implies that 
	\begin{align*}
	\abs{\Risk_{L^*,\P}(f_{L^*,\P,\lambda_n}) - \Risk_{L^*, \P}(f_{L^*,\Dr,\lambda_n})} \leq \epsilon
	\end{align*}
	holds with probability tending to $1$. Since $\lambda_n \downarrow 0$, we additionally have, for all sufficiently large $n \in \N$, that
	\begin{align*}
	\abs{\Risk_{L^*, \P}(f_{L^*, \P, \lambda_n}) - \Risk^*_{L^*, \P}} \leq \epsilon
	\end{align*}
	and hence the assertion of $L^*$-risk consistency of $f_{L^*, \P, \lambda}$. \\
	
	In order to show the second assertion, we define for $n \in \N$, $\epsilon_n := (\ln(n+1))^{-1/2}$ and 
	\begin{align*}
	\delta_n := \Risk_{L^*, \P}(f_{L^*, \P, \lambda_n}) - \Risk^*_{L^*, \P} + \epsilon_n.
	\end{align*}
	For an infinite sample $D_\infty := ((X_i,Y_i,X'_i,Y'_i))_{i \in \N} \in ((\X \times \Y)^2)^\infty$ set 
	\begin{align*}
	D_n := ((X_i,Y_i,X'_i,Y'_i), \dots, (X_n,Y_n,X'_n,Y'_n)).
	\end{align*}
	We define for $n \in \N$
	\begin{align*}
	A_n := \{D_\infty \in ((\X \times \Y)^2)^\infty : \Risk_{L^*, \P}(f_{L^*, \Dr_n, \lambda_n}) - \Risk^*_{L^*, \P} > \delta_n\}.
	\end{align*}
	Now, our estimates above together with $\lambda_n^{2+\delta}n \to \infty$ for some $\delta > 0$ yield
	\begin{align*}
	\sum_{n \in \N} \P(A_n) \leq \sum_{n \in \N} \exp\Arg{-\frac{3}{8} \cdot \frac{\epsilon_n^2\lambda_n^2n}{(\epsilon_n\lambda_n + 3\abs{L}_1)\abs{L}_1}} < \infty.
	\end{align*}
	We obtain by the Borel-Cantelli lemma that
	\begin{align*}
	\P\Arg{\{D_\infty \in ((\X \times \Y)^2)^\infty: \Exists n_0 \in \N \, \Forall n \geq n_0 \text{ with } \Risk_{L^*, \P}(f_{L^*,\Dr_n, \lambda_n})\}} = 1.
	\end{align*}
	The assertion follows as $\lambda_n \to 0$ implies $\delta_n \to 0$.
\end{Proof}

\begin{lemma}\label{lem:regRisk0}
	Let $f_{L^*,\P,\lambda} \in \H$ be any fixed minimizer of $\inf_{f \in \H} \left(\Risk_{L^*,\P}(f) + \lambda\norm{f}_\H^2\right)$. Then we have, for any $g \in \H$,
	\begin{align*}
		\E{D_5 L(x,y,x',y',f_{L^*,\P,\lambda}(x,x'))g(x,x')}{\P^2} + 2\lambda\left<f_{L^*,\P,\lambda},g\right>_\H = 0.
	\end{align*}
\end{lemma}

\begin{Proof}{\Cref{lem:regRisk0}}
	Abbreviate $f_\P := f_{L^*,\P,\lambda}$, as we consider $L^*$ and $\lambda$ to be fixed in this proof. Let $g \in \H$. We define $$\tilde{G}: [-1,1] \to \R, \qquad\tilde{G}(t) = \Risk_{L^*,\P}(f_\P + tg) + \lambda\norm{f_\P + tg}_\H^2.$$
	$\tilde{G}$ is continuous as it is a composition of continuous functions. Recall that the derivatives of $L$ and $L^*$ with respect to the fifth argument are identical because $L$ and $L^*$ only differ by the term $L(x,y,x',y',0)$. For $t \ne 0$, we obtain by using the Lipschitz continuity of $L$, that
	\begin{align}
	&\hspace*{1.3em} \frac{\tilde{G}(t) - \tilde{G}(0)}{t} \nonumber\\
	&= \frac{1}{t}\left(\Risk_{L^*,\P}(f_P + tg) + \lambda\norm{f_\P + tg}_\H^2 - \Risk_{L^*,\P}(f_\P) - \lambda\norm{f_\P}_\H^2\right) \nonumber\\
	&= \frac{1}{t}\int_{(\X \times \Y)^2} L\bigl(x,y,x',y',f_\P(x,x')+tg(x,x')\bigr) - L\bigl(x,y,x',y',f_\P(x,x')\bigr) \diff \P^2(x,y,x',y') \nonumber\\
	&\quad + \frac{1}{t}\lambda\left<f_\P+tg,f_\P+tg\right>_\H - \frac{1}{t}\lambda\norm{f_\P}_\H^2 \\
	&= \frac{1}{t}\int_{(\X \times \Y)^2} L\bigl(x,y,x',y',f_\P(x,x')+tg(x,x')\bigr) - L\bigl(x,y,x',y',f_\P(x,x')\bigr) \diff \P^2(x,y,x',y') \nonumber\\
	&\quad + 2\lambda\left<f_\P,g\right>_\H + t\norm{g}_\H^2 \nonumber\\
	&\leq \frac{1}{t}\int_{(\X \times \Y)^2}\abs{L}_1\abs{tg(x,x')}\diff \P^2(x,y,x',y') +  2\lambda\left<f_\P,g\right>_\H + t\norm{g}_\H^2 \nonumber\\
	&= \abs{L}_1 \int_{(\X \times \Y)^2} \abs{g(x,x')} \diff \P^2(x,y,x',y') +  2\lambda\left<f_\P,g\right>_\H + t\norm{g}_\H^2 \nonumber\\
	&\leq \abs{L}_1\norm{g}_\infty +  2\lambda\left<f_\P,g\right>_\H + t\norm{g}_\H^2 \nonumber\\
	&< \infty \nonumber.
	\end{align} 
	Furthermore, we have for all $(x,y,x',y') \in (\X \times \Y)^2$,
	\begin{align*}
	\lim_{t \to 0} \frac{1}{t}\Bigl(L\bigl(x,y,x',y',f_\P(x,x')+tg(x,x')\bigr) - L\bigl(x,y,x',y',f_\P(x,x')\bigr)\Bigr) = D_5 L(x,y,x',y',f_\P(x,x'))g(x,x').
	\end{align*}
	Therefore (A.3) and an application of Lebesgue's theorem of dominated convergence yield
	\begin{align*}
	\lim_{t \to 0} \frac{\tilde{G}(t) - \tilde{G}(0)}{t} &= \int_{(\X \times \Y)^2} D_5 L(x,y,x',y',f_\P(x,x'))g(x,x') \diff \P^2(x,y,x',y') + 2\lambda\left<f_\P,g\right>_\H.
	\end{align*}
	We know from \Cref{Inequalities}(iii) that $$\tilde{G}(0) = \Risk^\text{reg}_{L^*, \P, \lambda}(f_\P) = \inf_{f \in \H} \Risk^\text{reg}_{L^*,\P, \lambda}  \leq 0$$ and therefore $\tilde{G}(t) \geq \tilde{G}(0)$ which yields $\tilde{G}(t) - \tilde{G}(0) \geq 0$. This inequality also holds for the function $-g$. Hence the desired identity follows.
\end{Proof}

\begin{theorem}\label{contDiff}
	The function $G: \R \times \H \to \H$ defined by
	$$G(\epsilon, f) := 2\lambda f + \E{D_5 L(x,y,x',y',f(x,x'))\Phi(x,x')}{\P_\epsilon^2}$$ 
	with $\P_\epsilon = (1-\epsilon)\P + \epsilon \Qp$ is continuously differentiable and $\frac{\partial G}{\partial f}(0,f)$ is invertible for all $f \in \H$.
\end{theorem}

\begin{Proof}{\Cref{contDiff}}
	We use \Cref{partialFrechetDifferentiability} and will show that $\frac{\partial G}{\partial \epsilon}$ and $\frac{\partial G}{\partial f}$ are continuous. To shorten the notation in the proof, set $L'_f(X,Y,X',Y') := D_5 L(X,Y,X',Y',f(X,X'))$ and $L''_f(X,Y,X',Y') := D_5L'_f(X,Y,X',Y')$. Note that for $\epsilon \in \R$ and $f \in \H$,
	\begin{align*}
	\frac{\partial G}{\partial \epsilon}(\epsilon, f) &= -2(1-\epsilon)\E{L'_f(X,Y,X',Y')\Phi(X,X')}{\P^2} + (1-2\epsilon)\E{L'_f(X,Y,X',Y')\Phi(X,X')}{\P \otimes \Qp}\\
	&\quad + (1-2\epsilon)\E{L'_f(X,Y,X',Y')\Phi(X,X')}{\Qp \otimes \P} + 2\epsilon\E{L'_f(X,Y,X',Y')\Phi(X,X')}{\Qp^2}.
	\end{align*}
	For $\epsilon, \tilde{\epsilon} \in \R$ and $f, \tilde{f} \in \H$, we have 
	\begin{align*}
	\frac{\partial G}{\partial \epsilon}(\epsilon, f) - \frac{\partial G}{\partial \epsilon}(\tilde{\epsilon}, \tilde{f}) &= \left(\frac{\partial G}{\partial \epsilon}(\epsilon, f) - \frac{\partial G}{\partial \epsilon}(\epsilon, \tilde{f})\right) + \left(\frac{\partial G}{\partial \epsilon}(\epsilon, \tilde{f}) - \frac{\partial G}{\partial \epsilon}(\tilde{\epsilon}, \tilde{f})\right)\\
	&=: \partial G_1 + \partial G_2.
	\end{align*}
	Here $\partial G_1$ equals
	\begin{align*}
	& -2(1-\epsilon)\E{(L'_f-L'_{\tilde{f}})(X,Y,X',Y')\Phi(X,X')}{\P^2} + (1-2\epsilon)\E{(L'_f-L'_{\tilde{f}})(X,Y,X',Y')\Phi(X,X')}{\P \otimes \Qp}\\
	&+ (1-2\epsilon)\E{(L'_f-L'_{\tilde{f}})(X,Y,X',Y')\Phi(X,X')}{\Qp \otimes \P} + 2\epsilon\E{(L'_f-L'_{\tilde{f}})(X,Y,X',Y')\Phi(X,X')}{\Qp^2}.
	\end{align*}
	Set $Z := \{(x,y,x',y') \in (\X \times \Y)^2: f(x,x') \ne \tilde{f}(x,x')\}$. We compute the expectation with respect to probability measures $\P_1, \P_2 \in \M_1(\X \times \Y)$ first, in order to simplify the term above. An application of the mean value theorem (MVT) and the boundedness of the second derivative yield, for all $\P_1, \P_2 \in \M_1(\X \times \Y)$, that
	\begin{align*}
	&\quad \norm{\E{(L'_f - L'_{\tilde{f}})(X,Y,X',Y')\Phi(X,X')}{\P_1 \otimes \P_2}}_\H\\
	&\leq \E{\norm{(L'_f - L'_{\tilde{f}})(X,Y,X',Y')\Phi(X,X')}_\H}{\P_1 \otimes \P_2} = \E{\abs{(L'_f-L'_{\tilde{f}})(X,Y,X',Y')}\norm{\Phi(X,X')}_\H}{\P_1 \otimes \P_2}\\
	&= \int_{(\X \times \Y)^2} \abs{(L'_f-L'_{\tilde{f}})(x,y,x',y')}\norm{\Phi(x,x')}_\H \diff (\P_1 \otimes \P_2)(x,y,x',y')\\
	&= \int_Z \abs{f(x,x')-\tilde{f}(x,x')}\abs{\frac{(L'_f-L'_{\tilde{f}})(x,y,x',y')}{f(x,x')-\tilde{f}(x,x')}}\norm{\Phi(x,x')}_\H\diff (\P_1 \otimes \P_2)(x,y,x',y')\\
	&\leftstackrel{\text{MVT}}{\leq} \int_Z \abs{f(x,x')-\tilde{f}(x,x')}c_{L,2}\norm{\Phi(x,x')}_\H \diff (\P_1 \otimes \P_2)(x,y,x',y')\\
	&\leftstackrel{}{\leq} \int_Z \norm{f-\tilde{f}}_\infty c_ {L,2} \norm{k}_\infty	\diff (\P_1 \otimes \P_2)(x,y,x',y')\\
	&\leftstackrel{}{\leq}\int_Z \norm{f-\tilde{f}}_\H c_ {L,2} \norm{k}^2_\infty	\diff (\P_1 \otimes \P_2)(x,y,x',y')\\
	&= \norm{f-\tilde{f}}_\H c_ {L,2} \norm{k}^2_\infty < \infty.
	\end{align*}
	As this upper bound is valid for all $\P_1, \P_2 \in \M_1(\X \times \Y)$, the desired result for $\partial G_1$ follows. We have
	\begin{align*}
	\quad \norm{\partial G_1}_\H &\leq \left(2\abs{1-\epsilon} + 2\abs{1-2\epsilon} + 2\abs{\epsilon}\right) \cdot \norm{\E{(L'_f - L'_{\tilde{f}})(X,Y,X',Y')\Phi(X,X')}{\P_1 \otimes \P_2}}_\H\\
	&\leq (2(1+\abs{\epsilon}) + 2(1+2\abs{\epsilon}) + 2\abs{\epsilon}) \cdot \bigl(\norm{k}^2_\infty c_{L,2}\lvert\lvert f-\tilde{f}\rvert\rvert_\H\bigr)\\
	&= (4+8\abs{\epsilon}) \cdot \bigl(\norm{k}^2_\infty c_{L,2}\lvert\lvert f-\tilde{f}\rvert\rvert_\H\bigr).
	\end{align*}
	Moreover,
	\begin{align*}
	\partial G_2 &= 2(\epsilon-\tilde{\epsilon})\E{L'_{\tilde{f}}(X,Y,X',Y')\Phi(X,X')}{\P^2} + 2(\tilde{\epsilon}-\epsilon)\E{L'_{\tilde{f}}(X,Y,X',Y')\Phi(X,X')}{\P \otimes \Qp} \\
	&\quad + 2(\tilde{\epsilon} - \epsilon)\E{L'_{\tilde{f}}(X,Y,X',Y')\Phi(X,X')}{\Qp \otimes \P} + 2(\epsilon - \tilde{\epsilon})\E{L'_{\tilde{f}}(X,Y,X',Y')\Phi(X,X')}{\Qp^2}.
	\end{align*}
	Hence, we have by the boundedness of the first derivative, using the same approach  as above
	\begin{align*}
	\norm{\partial G_2}_\H &\leq 8\abs{\epsilon - \tilde{\epsilon}}\E{\norm{L'_{\tilde{f}}(X,Y,X',Y')\Phi(X,X')}_\H}{\P_1 \otimes \P_2}\\
	&= 8\abs{\epsilon - \tilde{\epsilon}}\E{\abs{L'_{\tilde{f}}(X,Y,X',Y')}\norm{\Phi(X,X')}_\H}{\P_1 \otimes \P_2}\\
	&= 8\abs{\epsilon - \tilde{\epsilon}}c_{L,1}\norm{k}_\infty.
	\end{align*}
	Thus
	\begin{align*}
	\norm{\frac{\partial G}{\partial \epsilon}(\epsilon, f) - \frac{\partial G}{\partial \epsilon}(\tilde{\epsilon}, \tilde{f}) }_\H \leq (4+8\abs{\epsilon}) \cdot \left(\norm{k}^2_\infty c_{L,2}\lvert\lvert f-\tilde{f} \rvert\rvert_\H\right) + 8\abs{\epsilon - \tilde{\epsilon}}c_{L,1}\norm{k}_\infty.
	\end{align*}
	From this, we obtain for $\epsilon \to \tilde{\epsilon}$ and $f \to \tilde{f}$ the continuity of the partial derivative $\frac{\partial G}{\partial \epsilon}$. 
	
	The partial derivative $\frac{\partial G}{\partial \H}$ can be expressed as
	\begin{align*}
	\frac{\partial G}{\partial \H}(\epsilon, f) = 2\lambda \id_\H + \E{D_5L'_f(X,Y,X',Y') \left<\Phi(X,X'), \cdot\right>_\H \Phi(X,X')}{\P_\epsilon^2}, \quad \epsilon \in \R, f \in \H.
	\end{align*}
	To prove its continuity, we first observe, for any $\tilde{f} \in \H$,
	\begin{align*}
	&\quad \frac{\partial G}{\partial \H}(\epsilon, f) - \frac{\partial G}{\partial \H}(\epsilon, \tilde{f}) \\
	&= \E{D_5(L'_f(X,Y,X',Y'))\left<\Phi(X,X'), \cdot\right>_\H \Phi(X,X') - D_5(L'_{\tilde{f}}(X,Y,X',Y'))\left<\Phi(X,X'),\cdot\right>_\H\Phi(X,X')}{\P_\epsilon^2}\\
	&=\E{D_5(L'_f(X,Y,X',Y')-L'_{\tilde{f}}(X,Y,X',Y'))\left<\Phi(X,X'),\cdot\right>_\H\Phi(X,X')}{\P_\epsilon^2}.
	\end{align*}
	By the definition of the local modulus of continuity for the second order derivatives of $L$, see \Cref{localModulusOfContinuity}, and the Cauchy-Schwarz inequality, we obtain for $(x,x') \in \X^2$, that
	\begin{align*}
	\norm{\left<\Phi(x, x'), \cdot\right>\Phi(x, x')}_{\L(\H,\H)} &= \sup_{\substack{h \in \H\\\norm{h}_\H \leq 1}} \norm{\left<\Phi(x, x'), h\right>_\H\Phi(x, x')}_\H\\
	&= \sup_{\substack{h \in \H\\\norm{h}_\H \leq 1}} \abs{\left<\Phi(x, x'), h\right>_\H}\norm{\Phi(x, x')}_\H\\
	&\leq \sup_{\substack{h \in \H\\\norm{h}_\H \leq 1}} \norm{\Phi(x, x')}_\H \norm{h}_\H \norm{\Phi(x, x')}_\H\\
	&=  \norm{\Phi(x, x')}_\H^2 \leq \norm{k}_\infty^2.
	\end{align*}
	Hence, for $f, \tilde{f} \in \{g \in \H: \norm{g}_\H \leq r\}$, we obtain the upper bound
	\begin{align*}
	&\quad \norm{\frac{\partial G}{\partial \H}(\epsilon, f) - \frac{\partial G}{\partial \H}(\epsilon, \tilde{f})}_{\L(\H, \H)}\\
	&= \norm{\E{D_5(L'_f(X,Y,X',Y')-L'_{\tilde{f}}(X,Y,X',Y'))\left<\Phi(X,X'),\cdot\right>_\H\Phi(X,X')}{\P_\epsilon^2}}_{\L(\H, \H)}\\
	&\leq \E{\norm{D_5(L'_f(X,Y,X',Y')-L'_{\tilde{f}}(X,Y,X',Y'))\left<\Phi(X,X'),\cdot \right>_\H\Phi(X,X')}_{\L(\H, \H)}}{\P_\epsilon^2}\\
	&= \E{\abs{D_5(L'_f(X,Y,X',Y')-L'_{\tilde{f}}(X,Y,X',Y'))} \norm{\left<\Phi(X, X'), \cdot\right>_\H\Phi(X,X')}_{\L(\H,\H)}}{\P_\epsilon^2}\\
	&\leq \norm{k}_\infty^2 \E{\abs{D_5(L'_f(X,Y,X',Y')-L'_{\tilde{f}}(X,Y,X',Y'))}}{\P_\epsilon^2}\\
	&\leq \norm{k}_\infty^2 \omega\Arg{\norm{k}_\infty \norm{f-\tilde{f}}_\H}_{r\norm{k}_\infty}.
	\end{align*}
	The second difference of partial derivatives we need to consider is the following, in which the integrands are the same but the probability measures differ. We denote by $T_{\tilde{f}}(x,y,x',y')$ the derivative $D_5L'_{\tilde{f}}(x,y,x',y')\left<\Phi(x,x'), \cdot\right>_\H \Phi(x,x')$. We then obtain by elementary calculation,
	\begin{align*}
	&\quad \frac{\partial G}{\partial H}(\epsilon, \tilde{f}) - \frac{\partial G}{\partial H}(\tilde{\epsilon}, \tilde{f})\\
	 &= \E{D_5L'_{\tilde{f}}(X,Y,X',Y')\left<\Phi(X,X'), \cdot\right>_\H \Phi(X,X')}{\P_\epsilon^2} - \E{D_5L'_{\tilde{f}}(X,Y,X',Y')\left<\Phi(X,X'), \cdot\right>_\H \Phi(X,X')}{\P_{\tilde{\epsilon}}^2}\\
	&= (1-\epsilon)^2\E{T_{\tilde{f}}(X,Y,X',Y')}{\P^2} + (1-\epsilon)\epsilon\E{T_{\tilde{f}}(X,Y,X',Y')}{\P \otimes \Qp} + \epsilon(1-\epsilon)\E{T_{\tilde{f}}(X,Y,X',Y')}{\Qp \otimes \P} + \\
	&\quad \epsilon^2\E{T_{\tilde{f}}(X,Y,X',Y')}{\Qp^2} - (1-\tilde{\epsilon})^2\E{T_{\tilde{f}}(X,Y,X',Y')}{\P^2} - (1-\tilde{\epsilon})\tilde{\epsilon}\E{T_{\tilde{f}}(X,Y,X',Y')}{\P \otimes \Qp} - \\
	&\quad \tilde{\epsilon}(1-\tilde{\epsilon})\E{T_{\tilde{f}}(X,Y,X',Y')}{\Qp \otimes \P} - \tilde{\epsilon}^2\E{T_{\tilde{f}}(X,Y,X',Y')}{\Qp^2}\\
	&= (\tilde{\epsilon} - \epsilon)(2-\tilde{\epsilon} - \epsilon)\E{T_{\tilde{f}}(X,Y,X',Y')}{\P^2} + (\epsilon - \tilde{\epsilon})(1-\epsilon-\tilde{\epsilon})\E{T_{\tilde{f}}(X,Y,X',Y')}{\P \otimes \Qp} +\\
	&\quad (\epsilon - \tilde{\epsilon})(1- \epsilon - \tilde{\epsilon})\E{T_{\tilde{f}}(X,Y,X',Y')}{\Qp \times \P} + (\epsilon - \tilde{\epsilon})(\epsilon + \tilde{\epsilon})\E{T_{\tilde{f}}(X,Y,X',Y')}{\Qp^2}.
	\end{align*}
	Due to the boundedness of the second derivative and the inequality $\norm{\left<\Phi(x,x'), \cdot\right>_\H\Phi(x,x')}_{\L(\H,\H)} \leq \norm{k}_\infty^2$ for all $(x,x') \in \X^2$, it follows that
	\begin{align*}
	\norm{\frac{\partial G}{\partial H}(\epsilon, \tilde{f}) - \frac{\partial G}{\partial H}(\tilde{\epsilon}, \tilde{f})}_{\L(\H, \H)} &\leq c_{L,2}\norm{k}_\infty^2\abs{\epsilon - \tilde{\epsilon}}(4+4\abs{\epsilon} + 4\abs{\tilde{\epsilon}})\\
	&= 4c_{L,2}\norm{k}_\infty^2\abs{\epsilon - \tilde{\epsilon}}(1+\abs{\epsilon}+\abs{\tilde{\epsilon}}).
	\end{align*}
	Hence
	\begin{align*}
	\norm{\frac{\partial G}{\partial H}(\epsilon, f) - \frac{\partial G}{\partial H}(\tilde{\epsilon}, \tilde{f})}_{\L(\H, \H)} &= 	\norm{\frac{\partial G}{\partial H}(\epsilon, f) - \frac{\partial G}{\partial H}(\epsilon, \tilde{f}) + \frac{\partial G}{\partial H}(\epsilon, \tilde{f}) - \frac{\partial G}{\partial H}(\tilde{\epsilon}, \tilde{f})}_{\L(\H, \H)}\\
	&\leq \norm{\frac{\partial G}{\partial H}(\epsilon, \tilde{f}) - \frac{\partial G}{\partial H}(\tilde{\epsilon}, \tilde{f})}_{\L(\H, \H)} + \norm{\frac{\partial G}{\partial \H}(\epsilon, f) - \frac{\partial G}{\partial \H}(\epsilon, \tilde{f})}_{\L(\H, \H)}\\
	&\leq \norm{k}_\infty^2 \omega\Bigl(\norm{k}_\infty \bigl\lvert\bigl\lvert f-\tilde{f} \bigr\rvert\bigr\rvert_\H\Bigr)_{r\norm{k}_\infty} + 4c_{L,2}\norm{k}_\infty^2\abs{\epsilon - \tilde{\epsilon}}(1+\abs{\epsilon}+\abs{\tilde{\epsilon}}),
	\end{align*}
	which yields the continuity of the partial derivative $\frac{\partial G}{\partial H}$ and thus the continuous differentiability of $G$. Let $f \in \H$ and consider the linear operator $\frac{\partial G}{\partial H}(0,f)$. We obtain
	\begin{align*}
	\frac{\partial G}{\partial H}(0,f) &= 2\lambda\id_\H + \E{D_5L'_f(X,Y,X',Y')\left<\Phi(X,X'), \cdot\right>_\H \Phi(X,X')}{\P^2}.
	\end{align*}
	Hence, for all $g, \tilde{g} \in \H$,
	\begin{align*}
	\left<\frac{\partial G}{\partial H}(0,f)(g), \tilde{g}\right>_\H &= 2\lambda\left<g,\tilde{g}\right>_\H + \E{D_5L'_f(X,Y,X',Y')\left<\Phi(X,X'), g\right>_\H\left<\Phi(X,X'),\tilde{g}\right>_\H}{\P^2}\\
	&= 2\lambda\left<g,\tilde{g}\right>_\H + \E{D_5L'_f(X,Y,X',Y')g(X,X')\tilde{g}(X,X')}{\P^2}.
	\end{align*}
	Therefore, the linear operator $\left<\frac{\partial G}{\partial H}(0,f)(g), \tilde{g}\right>_\H$ is symmetric. Hence its spectrum lies in the closed interval $[a,b]$ where
	\begin{align*}
	a := \inf_{\norm{g}_\H = 1} \left<\frac{\partial G}{\partial H}(0,f)(g), g\right>_\H, \qquad b := \sup_{\norm{g}_\H = 1} \left<\frac{\partial G}{\partial H}(0,f)(g), g\right>_\H.
	\end{align*}
	Due to Assumption 4.2, $L$ is a convex loss function. This implies that the second derivative with respect to the fifth argument is non-negative. Hence, we obtain by the convexity of $L$
	\begin{align*}
	\left<\frac{\partial G}{\partial H}(0,f)(g), g\right>_\H &= 2\lambda\left<g,g\right>_\H + \E{D_5L'_f(X,Y,X',Y') \left<\Phi(X,X'),g\right>_\H \left<\Phi(X,X'),g\right>_\H}{\P^2}\\
	&= 2\lambda\norm{g}_\H^2 + \mathbb{E}_{\P^2}\Bigl[\underbrace{D_5L'_f(X,Y,X',Y')g^2(X,X')}_{\geq 0}\Bigr]\\
	&\geq 2\lambda\norm{g}_\H^2,
	\end{align*}
	for $g \in \H$. Thus it also applies for normalized functions, hence $a \geq 2\lambda > 0$. This shows that the operator $\frac{\partial G}{\partial H}(\epsilon, \tilde{f})(0,f)$ is invertible.
\end{Proof}

\begin{Proof}{\Cref{maxBias}}
	Denote with $T(x,y,x',y') := h_\P(x,y,x',y')\Phi(x,x')$. Using Fubini's theorem and the inequality 
	\begin{align*}
		\norm{T(x,y,x',y')}_\H = \norm{h(x,y,x',y')\Phi(x,x')}_\H &\leq \norm{h}_\infty \norm{\Phi(x,x')}_\H \stackrel{}{\leq} \abs{L}_1 \norm{k}_\infty,
	\end{align*} it follows by rearranging terms
	\begin{align*}
	&\quad \lambda\norm{f_{L^*,\P,\lambda} - f_{L^*,\P_\epsilon^2, \lambda}}_\H \\
	&\leq \norm{\E{h_\P\Phi}{\P^2} - \E{h_\P\Phi}{\P_\epsilon^2}}_\H\\
	&= \bigg\lVert\int\int T(x,y,x',y')\diff \P(x,y) \diff \P(x',y')- \int\int T(x,y,x',y')\diff \left[(1-\epsilon)\P + \epsilon \Qp\right] \diff \left[(1-\epsilon)\P + \epsilon \Qp\right]\bigg\rVert_\H\\
	&= \bigg\lVert \epsilon\int\int T(x,y,x',y')\diff \P(x,y)\diff (\P-\Qp)(x',y') + \epsilon\int\int T(x,y,x',y')\diff (\P-\Qp)(x,y)\diff \P(x',y')\\
	&\quad +\epsilon^2\int\int T(x,y,x',y')\diff \P(x,y)\diff(\Qp-\P)(x',y') + \epsilon^2\int\int T(x,y,x',y')\diff \Qp(x,y)\diff (\P-\Qp)(x',y')\bigg\rVert_\H\\
	&\leq 4\epsilon\abs{L}_1\norm{k}_\infty \underbrace{\norm{\P-\Qp}_{TV}}_{\leq 2}\\
	&\leq 8\epsilon\abs{L}_1\norm{k}_\infty,
	\end{align*}
	where $\norm{\P-\Qp}_{TV}$ denotes the norm of total variation, i.e.
	\begin{align*} 
		\norm{\P-\Qp}_{TV} := \sup_{\substack{\norm{g}_\infty \leq 1\\ g: (\X \times \Y)^2 \to \R}} \abs{\int g \diff\P - \int g \diff \Qp}.
	\end{align*}
It is well-known that $\norm{\P-\Qp}_{TV} \in [0,2]$ for all $\P, \Qp \in \M_1(\X \times \Y)$.
	In conclusion 
	\begin{align*}
	\norm{f_{L^*,P,\lambda} - f_{L^*,P_\epsilon,\lambda}}_\H &\leq \frac{8}{\lambda}\abs{L}_1\norm{k}_\infty \epsilon. \qedhere
	\end{align*}
\end{Proof}

\begin{Proof}{\Cref{RPLGateaux}}
	Fix $\Qp \in \M_1(\X \times \Y)$ and $\lambda \in (0, \infty)$. Denote $\P_\epsilon := (1-\epsilon)\P + \epsilon \Qp$ with $\epsilon \in (0,1)$. The function $G: \R \times \H,$ defined by $$G(\epsilon, f) = 2\lambda\id_\H + \E{D_5L(X,Y,X',Y',f(X,X'))\Phi(X,X')}{\P_\epsilon^2}$$ plays an important role in this proof. Since $k$ is bounded, all functions $f \in \H$ in the corresponding RKHS fulfill $\norm{f}_\infty < \infty$. Additionally the partial derivative $D_5L$ is bounded by \Cref{assumptions}. It follows, for all $\epsilon \in \R$, and all $f \in \H$, that 
	\begin{align*}
	\norm{G(\epsilon, f)}_\H &\leq 2\lambda\norm{f}_\H + \E{\abs{D_5L(X,Y,X',Y',f(X,X'))} \sup_{(x, x') \in \X^2} \abs{\Phi(x,x')}}{\P_\epsilon^2}\\
	&\leq 2\lambda\norm{f}_\H + c_{L,1}\norm{k}_\infty < \infty.
	\end{align*}	 
	Therefore, the map $G$ is well-defined and bounded with respect to the $\H$-norm. Hence,
	\begin{align*}
	\norm{G(\epsilon, f)}_\infty &\leq \norm{G(\epsilon, f)}_\H \norm{k}_\infty \leq (2\lambda\norm{f}_\H + c_{L,1}\norm{k}_\infty)\norm{k}_\infty < \infty.
	\end{align*}
	Note, that for $\epsilon \notin [0,1]$ the $\H$-valued Bochner integral is with respect to a signed measure. Hence \Cref{Risk-dbar} yields, for all $\epsilon \in [0,1]$, that
	\begin{align*}
	G(\epsilon, f) = \frac{\partial (\Risk_{L^*,\P_\epsilon}(\cdot) + \lambda\norm{\cdot}_\H)}{\partial H}(f).
	\end{align*}
	Since $L$ is convex, the map $f \mapsto \Risk_{L^*,\P_\epsilon}(f) + \lambda\norm{f}_\H^2$ is continuous and convex for all $\epsilon \in [0,1]$. The equation above shows that we have $G(\epsilon, f) = 0$ if and only if $f = f_{L^*, \P_\epsilon, \lambda}$ for such $\epsilon$. We now want to show the existence of a differentiable function $\epsilon \mapsto f_\epsilon$ on a small interval $(-\delta, \delta)$ for some $\delta > 0$ that satisfies $G(\epsilon, f_\epsilon) = 0$ for all $\epsilon \in (-\delta, \delta)$. According to the Implicit Function Theorem \ref{implicit-function-thm}, we have to check that $G$ is continuously differentiable and that $\frac{\partial G}{\partial \H}(0, f_{\P, \lambda})$ is invertible which was proven in \Cref{contDiff}.
	Hence we can apply the implicit function theorem to see that the map $\epsilon \mapsto f_\epsilon$ is differentiable on a small non-empty interval $(-\delta, \delta)$. In conclusion, we obtain
	\begin{align*}
	S'_G(\P)(\Qp) &= \frac{\partial f_\epsilon}{\partial \epsilon}(0) = -\left(\frac{\partial G}{\partial \H}(0, f_{L^*, \P, \lambda})\right)^{-1} \circ \frac{\partial G}{\partial \epsilon}(0, f_{L^*, \P, \lambda}) = -M(\P)^{-1}T(\Qp;\P),
	\end{align*}
	which yields the assertion.
\end{Proof}

\begin{Proof}{\Cref{boundedInfluenceFunction}}
	The assertion follows immediately by setting $\Qp$ as the Dirac measure $\delta_{(x_0, y_0)}$ in \Cref{RPLGateaux}. \qedhere
\end{Proof}

\begin{Proof}{\Cref{continuityOperator}}
	To (i). Let $\P \in \M_1(\X \times \Y)$ be fixed. As $L^*$ and $\lambda$ are fixed, we denote with
	\begin{align*}
	{L^*}'_{f_{L^*,\P,\lambda}}(X,Y,X',Y') \stackrel{\ref{Inequalities}\textrm{(v)}}{=} L'_{f_{L^*,\P,\lambda}}(X,Y,X',Y') := D_5L^*(X,Y,X',Y',f_{L^*, \P, \lambda}(X,X')).
	\end{align*}
	Let $(\P_n)_{n \in \N} \subset \M_1(\X \times \Y)$ be a weakly convergent sequence with $\P_n \rightsquigarrow \P$. We know that due to the separability of $\X \times \Y$, weak convergence of probability measures is equivalent to $d_{BL}(\P_n, \P) \to 0$, where $d_{BL}$ denotes the bounded Lipschitz metric, see \citet[Thm.~11.3.3]{dudley2002}.
	Hence the metric space $(\X \times \Y)^2$ is separable and thus guarantees
	\begin{align*}
	\P_n \rightsquigarrow \P \Longleftrightarrow \P_n^2 \rightsquigarrow \P^2 \quad (n \to \infty)
	\end{align*}
	see \citet[Thm.\,3.8(ii), p.\,23]{billingsley1999}.
	The definition of weak convergence guarantees that 
	\begin{align*}
	\lim_{n \to \infty} \int g \diff \P_n^2 = \int g \diff \P^2
	\end{align*} for all continuous and bounded real-valued functions $g: (\X \times \Y)^2 \to \R$. However, we need a corresponding result for $\H$-valued Bochner integrals. The fourth part of the representer theorem \ref{representerTheorem} yields
	\begin{align*}
	\norm{S(\P_n) - S(\P)}_\H &= \norm{f_{L^*, \P_n, \lambda} - f_{L^*, \P, \lambda}}_\H\\
	&\leq \frac{1}{\lambda}\norm{\E{h_\P(X,Y,X',Y')\Phi(X,X')}{\P_n^2} - \E{h_\P(X,Y,X',Y')\Phi(X,X')}{\P^2}}_\H.
	\end{align*}
	As $k$ is a continuous and bounded kernel, the canonical feature map $\Phi$ is also continuous and bounded.  Furthermore, as the shifted loss function $L^*$ is twice continuously differentiable and the partial derivative are bounded, it follows that, for every fixed $\P \in \M_1(\X \times \Y)$ and every fixed $\lambda \in (0, \infty)$, the function
	\begin{align*}
	&\psi_P: ((\X \times \Y)^2, d_{(\X \times \Y)^2}) \to (\H, d_\H), \quad \psi_\P(x,y,x',y') := h_\P(x,y,x',y')\Phi(x,x')
	\end{align*}
	is continuous and bounded, where $d_\H$ denotes the metric guaranteed by the norm $\norm{\cdot}_\H$. We thus obtain from \citet[p.\,III.40]{bourbaki2004}, see also \citet[Thm.\,A.1]{hable2011}, the following convergence result for Bochner integrals
	\begin{align*}
	\P_n^2 \rightsquigarrow \P^2 \Longrightarrow \lim_{n \to \infty} \int \psi_\P \diff \P_n^2 = \int \psi_\P \diff \P^2,
	\end{align*}
	which implies that $\P_n \rightsquigarrow \P$, which is equivalent to $d_*(\P_n, \P) \to 0$ due to \citet[Thm.\,11.3.3]{dudley2002}, leads to $\norm{S(\P_n) - S(\P)}_\H \to 0$ and therefore (i) is proven. \\
	
	The proof for (ii) follows immediately from part (i) and the fact that the inclusion map $\id: \H \to \Cb(\X^2)$ is continuous and bounded.
\end{Proof}

\begin{Proof}{\Cref{continuityEstimator}}
	Let $(D_{n})_{n \in \N} \subset (\X \times \Y)^n$ be a sequence which converges to some $D_{0} \in (\X \times \Y)^n$ for $n \to \infty$. Then the corresponding empirical measure $\D_{n}$ weakly converges to $\D_{0}$, i.e. $\D_{n} \rightsquigarrow \D_{0}$. Hence, the assertion follows from \Cref{continuityOperator} and $S(\D_n) = f_{L^*, \D_n, \lambda} = S_n(D_n)$.
\end{Proof}

\begin{Proof}{\Cref{qualitativeRobustness}}
	Fix $\lambda \in (0, \infty)$. For any $\{(x_1,y_1), \dots, (x_n,y_n)\} = D_n \in (\X \times \Y)^n$ denote its empirical measure by $\D_n := \frac{1}{n}\sum_{i=1}^n \delta_{(x_i,y_i)}$. According to \Cref{continuityEstimator}, the functions
	$$S_n: ((\X \times \Y)^n, d_{(\X \times \Y)^n}) \to (\H, d_\H), \quad S_n(D_n) = f_{L^*, \D_n, \lambda}$$
	are continuous and therefore measurable with respect to the corresponding Borel-$\sigma$-algebras for every $n \in \N$. \Cref{continuityOperator} yields that
	$$S: (\M_1(\X \times \Y), d_{BL}) \to (\H, d_\H), \quad S(\P) = f_{L^*, \P, \lambda}$$
	is a continuous operator. Furthermore $S_n$ and $S$ satisfy by definition the condition $S_n(D_n) = S(\D_n)$ for all $D_n \in (\X \times \Y)^n$ and all $n \in \N$. As $\H$ is a separable RKHS, $(\H, d_\H)$ is a complete and separable metric space. \Cref{thm:qualitativeRobustness} yields that for the random measure $\Dr_n = \frac{1}{n}\sum_{i=1}^n \delta_{(X_i,Y_i)}$ the sequence of RPL estimators $(f_{L^*,\Dr_n, \lambda})_{n \in \N}$ is qualitatively robust for all $\P \in \M_1(\X \times \Y)$. Hence the assertion of part (i) is shown.\\
	
	Part (ii) can be proven as follows. \Cref{continuityOperator} yields that the operator $S$ is continuous for all $\P \in \M_1(\X \times \Y)$. Hence all assumptions for \Cref{thm:bootstrapRobustness} are satisfied, because $\Zc := \X \times \Y$ is a compact metric space by assumption and $\W := \H$ is a complete and separable metric space. This yields the assertion.
\end{Proof}

\newpage

\addcontentsline{toc}{section}{References}
\nocite{*}
\printbibliography

\end{document}